\pdfoutput=1

\documentclass[11pt]{article}

\usepackage[]{acl}

\usepackage{times}
\usepackage{latexsym}

\usepackage[T1]{fontenc}

\usepackage[utf8]{inputenc}

\usepackage{microtype}

\usepackage{amsmath}
\usepackage{amsfonts}
\usepackage{amsthm}
\usepackage{amssymb}
\usepackage{graphicx}
\usepackage[ruled,linesnumbered]{algorithm2e} 
\usepackage{booktabs}
\usepackage{stackengine}

\DeclareMathOperator{\argmax}{argmax}

\DeclareMathOperator{\softmax}{Softmax}

\title{Revisiting Softmax for Uncertainty Approximation in Text Classification}
\author{
Andreas Nugaard Holm\and
Dustin Wright\and
Isabelle Augenstein\\
Department of Computer Science, University of Copenhagen\\
\{aholm, dw, augenstein\}@di.ku.dk
}

\begin{document}
\maketitle
\begin{abstract}
Uncertainty approximation in text classification is an important area with applications in domain adaptation and interpretability. One of the most widely used uncertainty approximation methods is Monte Carlo (MC) Dropout, which is computationally expensive as it requires multiple forward passes through the model. A cheaper alternative is to simply use the softmax based on a single forward pass without dropout to estimate model uncertainty. However, prior work has indicated that these predictions tend to be overconfident. In this paper, we perform a thorough empirical analysis of these methods on five datasets with two base neural architectures in order to identify the trade-offs between the two. We compare both softmax and an efficient version of MC Dropout on their uncertainty approximations and downstream text classification performance, while weighing their runtime (cost) against performance (benefit). We find that, while MC dropout produces the best uncertainty approximations, using a simple softmax leads to competitive and in some cases better uncertainty estimation for text classification at a much lower computational cost, suggesting that softmax can in fact be a sufficient uncertainty estimate when computational resources are a concern.
\end{abstract}

\section{Introduction}
The pursuit of pushing state-of-the-art performance on machine learning benchmarks often comes with an added cost of computational complexity. On top of already complex base models, such as Transformer models \cite{vaswani2017attention,lin2021survey}, successful methods often employ additional techniques to improve the uncertainty estimation of these models as they tend to be over-confident in their predictions. Though these techniques can be effective, the overall benefit in relation to the added computational cost is under-studied.

More complexity does not always imply better performance. For example, Transformers can be outperformed by much simpler convolutional neural nets (CNNs) when the latter are pre-trained as well \cite{tay-etal-2021-pretrained}.
Here, we turn our attention to neural network uncertainty estimation methods in text classification, which have applications in domain adaptation and decision making, and can help make models more transparent and explainable. In particular, we focus on a setting where efficiency is of concern, which can help improve the sustainability and democratisation of machine learning, as well as enable use in resource-constrained environments.

Quantifying predictive uncertainty in neural nets has been explored using various techniques~\cite{gawlikowski_survey_2021}, with the methods being divided into three main categories: Bayesian methods, single deterministic networks, and ensemble methods. Bayesian methods include Monte Carlo (MC) dropout~\cite{gal_dropout_2016} and Bayes by back-prop~\cite{blundell_weight_2015}. Single deterministic networks can approximate the predictive uncertainty by a single forward pass in the model, with softmax being the prototypical method. Lastly, ensemble methods utilise a collection of models to calculate the predictive uncertainty. However, while uncertainty estimation can improve when using more complex Bayesian and ensembling techniques, efficiency takes a hit.

In this paper, we perform an empirical investigation of the trade-off between choosing cheap vs. expensive uncertainty approximation methods for text classification, with the goal of highlighting the efficacy of these methods in an efficient setting.  We focus on one single deterministic and one Bayesian method. For the single deterministic method, we study the softmax, which is calculated from a single forward pass and is computationally very efficient. While softmax is a widely used method, prior work has posited that the softmax output when taken as a single deterministic operation is not the most dependable uncertainty approximation method~\cite{gal_dropout_2016,hendrycks_baseline_2018}. As such, it has been superseded by newer methods such as MC dropout, which leverages the dropout function in neural nets to approximate a random sample of multiple networks and aggregates the softmax outputs of this sample. MC dropout is favoured due to its close approximation of uncertainty, and because it can be used without any modification to the applied model. It has also been widely applied in text classification tasks~\cite{zhang_mitigating_2019,he-etal-2020-towards}.

To understand the cost vs. benefit of softmax vs. MC dropout, we perform experiments on five datasets using two different neural network architectures, applying them to three different downstream text classification tasks.
We measure both the added computational complexity in the form of runtime (cost) and the downstream performance on multiple uncertainty metrics (benefit).
We show that by using a single deterministic method like softmax instead of MC dropout, we can improve the runtime by $10$ times while still providing reasonable uncertainty estimates on the studied tasks.
As such, given the already high computational cost of deep neural network based methods and recent pushes for more sustainable ML \cite{strubell-etal-2019-energy,patterson2021carbon}, we recommend not discarding efficient uncertainty approximation methods such as softmax in resource-constrained settings, as they can still potentially provide reasonable estimations of uncertainty.

\textbf{Contribution} In summary, our contributions are:
1) an empirical study of an efficient version of MC dropout and softmax for text classification tasks, using two different neural architectures, and five datasets;
2) a comparison of uncertainty estimation between MC dropout and softmax using expected calibration error;
3) a comparison of the cost vs. benefit of MC dropout and softmax in a setting where efficiency is of concern.

\section{Related Work}
\subsection{Uncertainty Quantification}
Quantifying the uncertainty of a prediction can be done using various techniques~\cite{ovadia_can_2019,gawlikowski_survey_2021,conf/aaai/HenneSRW20} such as single deterministic methods~\cite{mozejko2019inhibited,vanamersfoort2020uncertainty} which calculate the uncertainty on a single forward pass of the model. They can further be classified as internal or external methods, which describe if the uncertainty is calculated internally in the model or post-processing the output. Another family of techniques are Bayesian methods, which combine NNs and Bayesian learning. Bayesian Neural Networks (BNNs) can also be split into subcategories, namely Variational Inference~\cite{hinton_keeping_1993}, Sampling~\cite{neal_bayesian_1993}, and Laplace Approximation~\cite{mackay_practical_1992}. Some of the more notable methods are Bayes by backprop~\cite{blundell_weight_2015} and Monte Carlo Dropout~\cite{gal_dropout_2016}. One can also approximate uncertainty using ensemble methods, which use multiple models to better measure predictive uncertainty, compared to using the predictive uncertainty given by a single model~\cite{conf/nips/Lakshminarayanan17,he-etal-2020-towards,durasov2021masksembles}.
Recently, we have seen uncertainty methods being used to develop methods for new tasks~\cite{zhang_mitigating_2019,he-etal-2020-towards}, where mainly Bayesian methods have been used.
We present a thorough empirical study of how uncertainty quantification behaves for text classification tasks. Unlike prior work, we do not only evaluate based on the performance of the methods, but perform an in-depth comparison to much simpler deterministic methods based on multiple metrics.

\subsection{Uncertainty Metrics}
Measuring the performance of uncertainty approximation methods can be done in multiple ways, each offering benefits and downsides. \citet{niculescu-mizil_predicting_2005} explore the use of obtaining confidence values from model predictions to use for supervised learning. One of the more widespread and accepted methods is using expected calibration error (ECE, \citealp{guo_calibration_2017}). While ECE measures the underlying confidence of the uncertainty approximation, we have also seen the use of human intervention for text classification~\cite{zhang_mitigating_2019,he-etal-2020-towards}. There, the uncertainty estimates are used to identify uncertain predictions from the model and ask humans to classify these predictions. The human classified data is assumed to have $100\%$ accuracy and to be suitable for measuring how well the model scores after removing a proportion of the most uncertain data points. Using metrics such as ECE, the calibration of models is shown, and this calibration can be improved using scaling techniques~\cite{guo_calibration_2017,naeini_obtaining_2015}.
We use uncertainty approximation metrics like expected calibration error, and human intervention (which we refer to as holdout experiments) to measure the difference in the performance of MC dropout and softmax compared against each other on text classification tasks.

\section{Uncertainty Approximation for Text Classification}
\begin{figure}[t]
\centering
\includegraphics[width=0.5\linewidth]{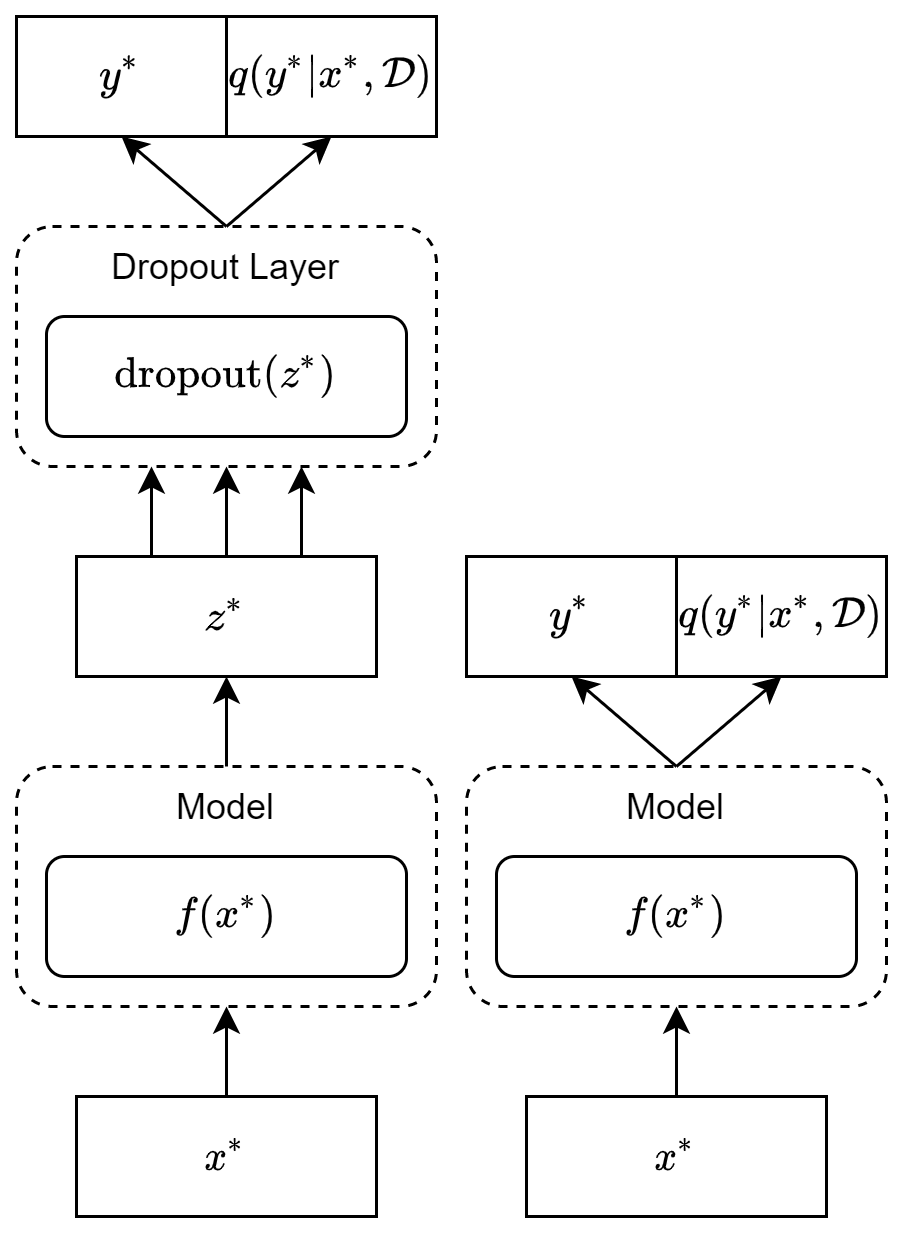}
\caption{MC Dropout (left) and softmax (right). In the version of MC dropout tested in this paper, a test input $x^{*}$ is passed through model $f$ to obtain a representation $z^{*}$, which is then subsequently passed through a dropout layer multiple times, and passed through the final part of the network to obtain prediction $y^{*}$. For softmax, dropout is disabled and a single prediction is obtained.}
\label{fig:mc_dropout}
\end{figure}

We focus on one deterministic method and one Bayesian method of uncertainty approximation. Both methods assume the existence of an already trained base model, and are applied at test time to obtain uncertainty estimates from the model's predictions. In the following sections, we formally introduce the two methods we study, namely MC dropout and softmax. MC dropout is a Bayesian method which utilises the dropout layers of the model to measure the predictive uncertainty, while softmax is a deterministic method that uses the classification output. In Figure \ref{fig:mc_dropout}, we visualise the differences between the two methods and how they are connected to base text classification models.

\subsection{Bayesian Learning}
\label{ssection:mcd}
Before introducing the MC dropout method, we quickly introduce the concept of \textit{Bayesian learning}. We start by comparing Bayesian learning to a traditional NN. A traditional NN assumes that the network weights $\omega \in \mathbb{R}^{n}$ are real but of an unknown value and can be found through maximum-likelihood estimation, and the input data $(x,y) \in \mathcal{D}$ are treated as random variables. Bayesian learning instead views the weights as random variables, and infers a posterior distribution $p(\omega|\mathcal{D})$ over $\omega$ after observing $\mathcal{D}$. The posterior distribution is defined as follows:
\begin{align}
    p(\omega|\mathcal{D})  &= \frac{p(\omega)p(\mathcal{D}|\omega)}{p(\mathcal{D})} = \frac{p(\omega)p(\mathcal{D}|\omega)}{\int p(\omega) p(\mathcal{D}|\omega)d\omega}
    \label{eq:bnn_true_posterior}
\end{align}
Using the posterior distribution, we can find the prediction of an input of unseen data $x^*$ and $y^*$ as follows:
\begin{align}
    p(y^*|x^*, \mathcal{D})  &= \int p(y^*|x^*, \omega) p(\omega|\mathcal{D}) d\omega.
    \label{eq:bnn_post}
\end{align}
However, the posterior distribution is infeasible to compute due to the marginal likelihood in the denominator, so we cannot find a solution analytically. We therefore resort to approximating the posterior distribution. For this approximation we rely on methods such as Bayes by Backpropagation~\cite{blundell_weight_2015} and Monte Carlo Dropout~\cite{gal_dropout_2016}.

\subsection{Monte Carlo Dropout}
At a high level, MC dropout approximates the posterior distribution $p(\omega | \mathcal{D})$ by leveraging the dropout layers in a model~\cite{gal_dropout_2016,gal_dropout_app_2016}. Mathematically, it is derived by introducing a distribution $q(\omega)$, representing a distribution of weight matrices whose columns are randomly set to $0$, to approximate the posterior distribution $p(\omega | \mathcal{D})$, which results in the following predictive distribution:
\begin{align}
    q(y^* \ | \ x^*, \mathcal{D})    &= \int p(y^*|x^*, \omega) q(\omega) d\omega.
    \label{eq:mc_post}
\end{align}
As this integral is still intractable, it is approximated by taking $K$ samples from $q(\omega)$ using the dropout layers of a learned network $f$ which approximates $p(y^*|x^*, \omega)$. As such, calculating $p(y^*|x^*, \omega) q(\omega)$ amounts to leaving the dropout layers active during testing, and approximating the integral amounts to aggregating predictions across multiple dropout samples. For the proofs, see~\citet{gal_dropout_2016}.

MC dropout requires multiple forward passes, so its computational cost is a multiple of the cost of performing a forward pass through the entire network. As this is obviously more computationally expensive than the single forward pass required for deterministic methods, we provide a fairer comparison between softmax and MC dropout by using an efficient version of MC dropout which caches an intermediate representation and only activates the dropout layers of the latter part of the network. As such, we obtain a representation $z^{*}$ by passing an input through the first several layers of the model and pass only this representation through the latter part of the model multiple times, reducing the computational cost while approximating the sampling of multiple networks.

\subsubsection{Combining Sample Predictions}
With multiple samples of the same data point, we have to determine how to combine them to quantify the predictive uncertainty. We test two methods that can be calculated using the logits of the model, requiring no model changes.
The first approach, which we refer to as Mean MC, is averaging the output of the softmax layer from all forward passes:
\begin{align}
    u_i = \frac{1}{K} \sum_{k=1}^K \softmax \left( f(z_i^k) \right),
\end{align}
where $z_i^k$ is a representation of the $i$'th data point of the $k$'th forward pass and $f$ is a fully-connected layer. The second method we use to quantify the predictive uncertainty is Dropout Entropy (DE)~\cite{zhang_mitigating_2019} which uses a combination of binning and entropy:
\begin{align}
    b_i &= \frac{1}{K}\text{BinCount}(\argmax(f(z_i))) \\
    u_i &= - \sum_{j=1}^C b_i(j) \log b_i(j)
\end{align}
$\text{BinCount}$ is the number of predictions of each class and $b$ is a vector the probabilities of a class's occurrence based on the bin count. We show the performance of the two methods in Section \ref{sssection:performance}.

\subsection{Softmax}
Softmax, a common normalising function for producing a probability distribution from neural network logits, is defined as follows:
\begin{align}
    u_i    &= \frac{e^{z_i}}{\sum^{C}_{j=1}e^{z_i(j)}},
\end{align}
where $z_i$ are the logits of the $i$'th data point. The softmax yields a probability distribution over the predicted classes. However, the predicted probability distribution is often overconfident toward the predicted class~\cite{gal_dropout_2016,hendrycks_baseline_2018}. The issue of softmax's overconfidence can also be exploited~\cite{gal_dropout_2016,joo_being_2020} -- in the worst case, this leads to the softmax producing imprecise uncertainties. However, model calibration methods like temperature scaling have been found to lessen the overconfidence to some extent~\cite{guo_calibration_2017}. As temperature scaling also incurs a cost in terms of runtime in order to find an optimal temperature, we choose to compare raw softmax probabilities to the efficient MC dropout method desribed previously, though uncertainty estimation could potentially be improved by scaling the logits appropriately.

\section{Experiments and Results}
\label{secction:experiments}
We consider five different datasets and two different base models in our experiments. Additionally, we conduct experiments to determine the optimal hyperparameters for the MC dropout method, particularly the optimal amount of samples which affects the efficiency and performance of MC dropout. In the paper we focus on the results of the 20 Newsgroups dataset, the results of the other four datasets are shown in the Appendix \ref{app:results} and \ref{app:model_calibration}. We further find the optimal dropout percentage in Appendix \ref{app:dropout}.

\subsection{Data}
To test the predictive uncertainty of the two methods, we use five datasets for diverse text classification tasks. We use the following five datasets:
The 20 Newsgroups dataset~\cite{lang_newsweeder_1995}, is a text classification consisting of a collection of $20.000$ news articles. The news articles are classified into 20 different classes.
The Amazon dataset~\cite{mcauley_hidden_2013} is a sentiment classification task. We use the `sports and outdoors' category, which consists of $272.630$ reviews ranging from $1$ to $5$.
The IMDb dataset~\cite{maas_learning_2011} is also a sentiment classification task. However, compared to the amazon dataset, this is a binary problem. The dataset consists of $50.000$ reviews.
The SST-2 dataset~\cite{socher_recursive_2013}, is also a binary sentiment classification dataset, consisting of $70.042$ sentences.
Lastly, we also use the WIKI dataset~\cite{redi_citation_2019}, which is a citation needed task, i.e. we predict if a citation is needed. The dataset consists of $19.998$ texts.
For the 20 Newsgroups, Amazon, IMDb and Wiki datasets, we use a split of $60$, $20$ and $20$ for the training, validation and test data, the data in splits have been selected randomly. We used the provided splits for the SST-2 dataset, but due to the test labels being hidden, we used the validation set for testing. We select these datasets as they are large, the tasks are diverse, and they cover multiple domains of text. Additionally, they represent well-studied and standard benchmarks in the field of text classification, which helps with the reproducibility of the results and comparison with baselines.

\subsection{Experimental Setup}
We use two different base neural architectures with two different embeddings in our experiments. To recreate baseline results, the first model is the same model as proposed in~\cite{zhang_mitigating_2019}, which is a CNN using pre-trained GloVe embeddings (Glove-CNN) with a dimension of $200$~\cite{pennington_glove_2014}. The second model uses a pre-trained BERT model~\cite{devlin_bert_2019} fine-tuned as masked language model on the dataset under evaluation to obtain contextualised embeddings, which are then input to a CNN with $4$ layers (BERT-CNN). The selection of these models allows us to compare the established baseline architecture from \cite{zhang_mitigating_2019} with a more modern version of it which takes advantage of large language models. For both models we use the final dropout layer for MC dropout.
Both models are optimised using Adam~\cite{kingma_adam_2015} and are trained for $1000$ epochs with early stopping after $10$ iterations if there have been no improvements, and we set the learning rate to $0.001$.

\paragraph{MC Dropout Sampling}
To make full use of MC dropout, we first determine the optimal number of forward passes through the model needed to obtain the best performance while maintaining high efficiency. This hyper-parameter search is imperative because the MC dropout performance and efficiency are correlated with the number of samples generated. To make a fair comparison against the already cheap softmax method, we want to find the minimum number of samples needed to approximate a good uncertainty. In Table \ref{table:sample_size}, we show the performance, using the F1 score, of the MC dropout method with the BERT-CNN model on the 20 Newsgroups dataset for the following number of samples: $\left[ 1, 5, 10, 25, 50, 100, 1000 \right]$. The table shows how the performance of the uncertainty approximation increases, given the number of samples. However, the performance gained by the number of samples falls off at 50. Given this, we use 50 MC samples in our experiments in order to balance good performance and efficiency.

\begin{table}[h]
\centering
\scalebox{0.85}{
\begin{tabular}{@{}lccccc@{}}
\toprule
1 & 10 & 25 & 50 & 100 & 1000 \\ \midrule
0.8212 & 0.8623 & 0.8540 & 0.8591 & 0.8559 & 0.8573 \\ \bottomrule
\end{tabular}
}
\caption{This table shows how the number of samples affect the performance of the MC dropout method, on the 20 Newsgroups dataset, using the BERT-CNN model. The results are reported using macro F1.}
\label{table:sample_size}
\end{table}

\subsection{Evaluation Metrics}
We use complementary evaluation metrics to benchmark the performance of MC dropout and softmax. Namely, we measure how well each of the methods identify uncertain predictions as well as the runtime of the methods.

\subsubsection{Efficiency}
To quantify efficiency, we measure the runtime of each of the methods during inference and the calculation of the uncertainties. Since we do not calculate uncertainties during training, this is only done on the test sets.
Training the model is independent of the uncertainty estimation methods, since we only use them to quantify the uncertainty of the predictions of the model. We therefore only calculate the runtime of each of the methods based on the test data.

\begin{figure*}[t]
    \begin{minipage}[t]{0.49\linewidth}
        \centering
        \includegraphics[width=1.0\linewidth]{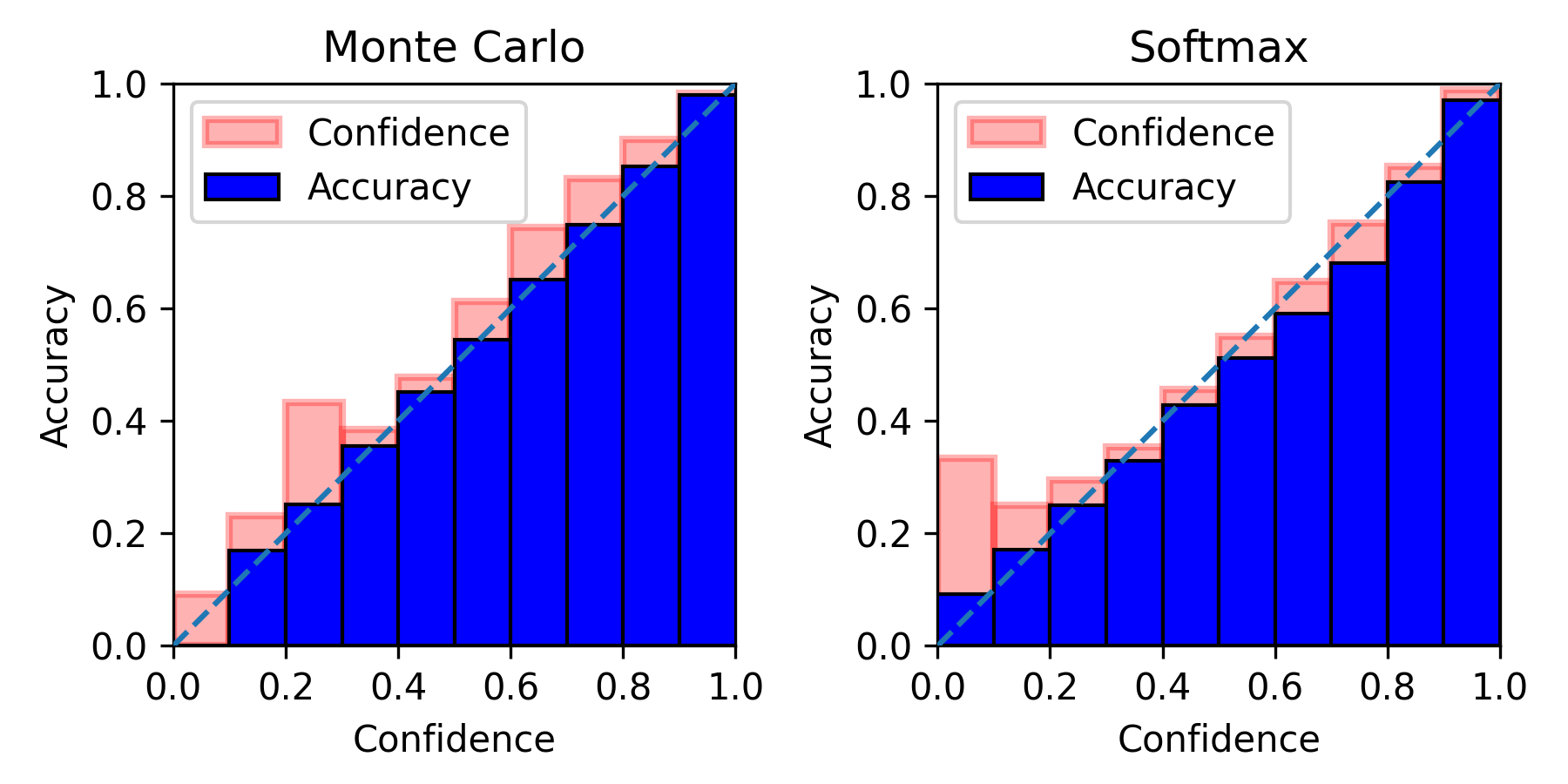}
    \end{minipage}
    \hfill
    \begin{minipage}[t]{0.49\linewidth}
        \centering
        \includegraphics[width=1.0\linewidth]{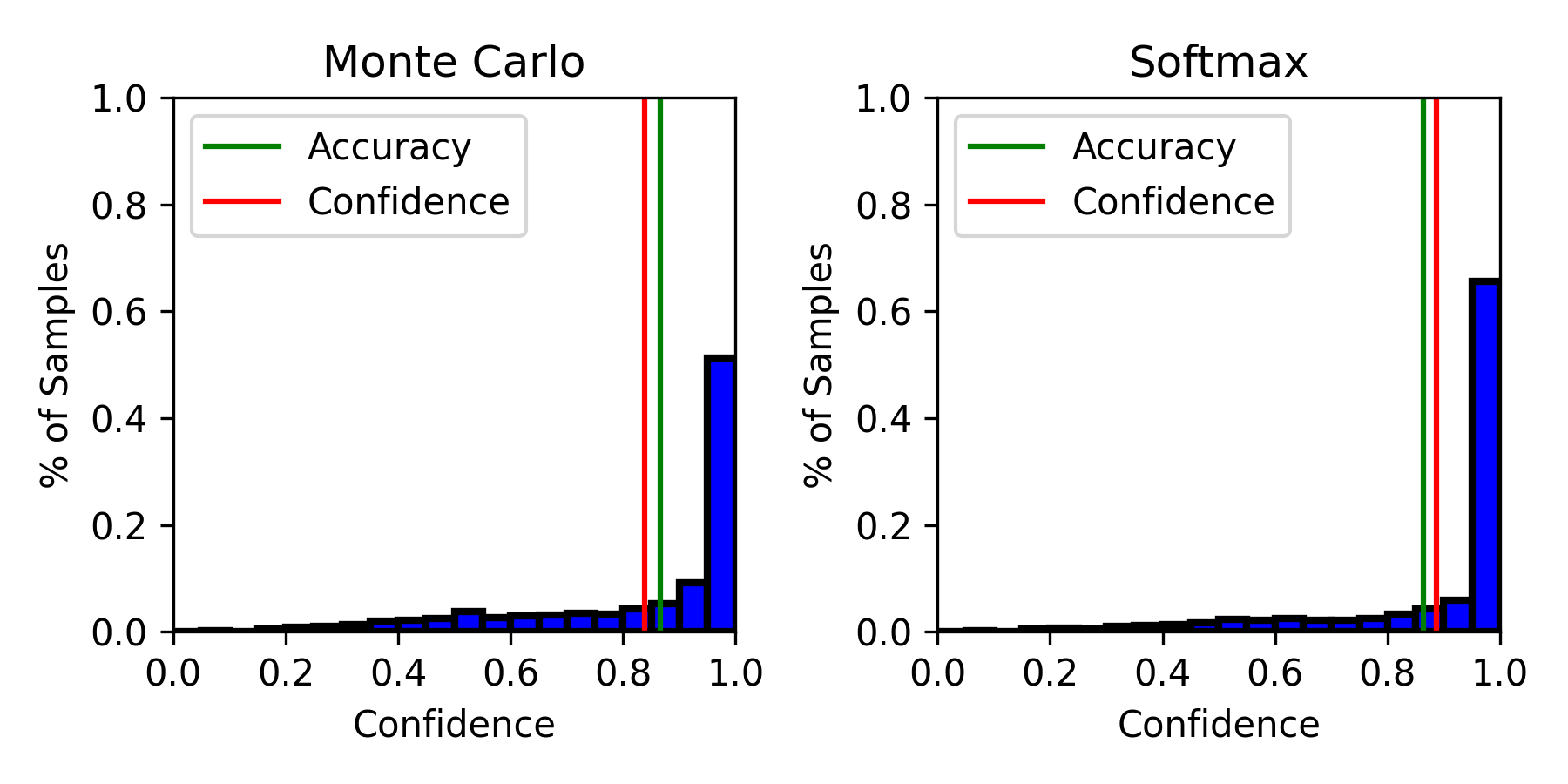}
    \end{minipage}
    \caption{Reliability diagram (left, displayed as a stacked bar chart comparing accuracy and confidence}) and confidence histogram (right) of 20 Newsgroups using BERT-CNN. Softmax and the efficient version of MC dropout tested in this paper are relatively similar in their calibration (a higher value for confidence than accuracy in any bin indicates overconfidence in that bin). At the same time, as indicated by the confidence histogram, softmax still produces more confident estimates on average.
    
    \label{fig:rel_conf_bert}
    \begin{minipage}[t]{0.49\linewidth}
        \centering
        \includegraphics[width=1.0\linewidth]{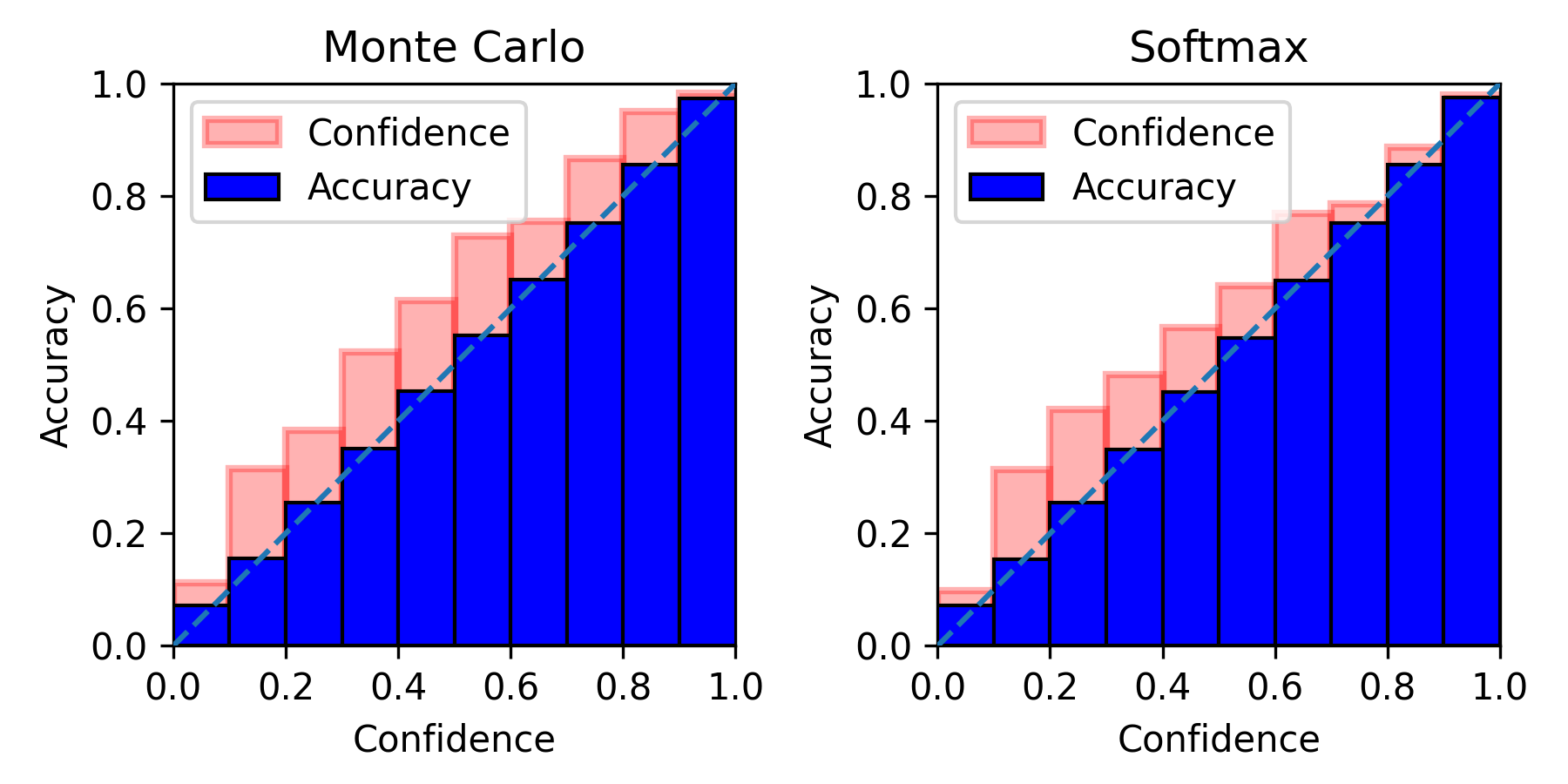}
    \end{minipage}
    \hfill
    \begin{minipage}[t]{0.49\linewidth}
        \centering
        \includegraphics[width=1.0\linewidth]{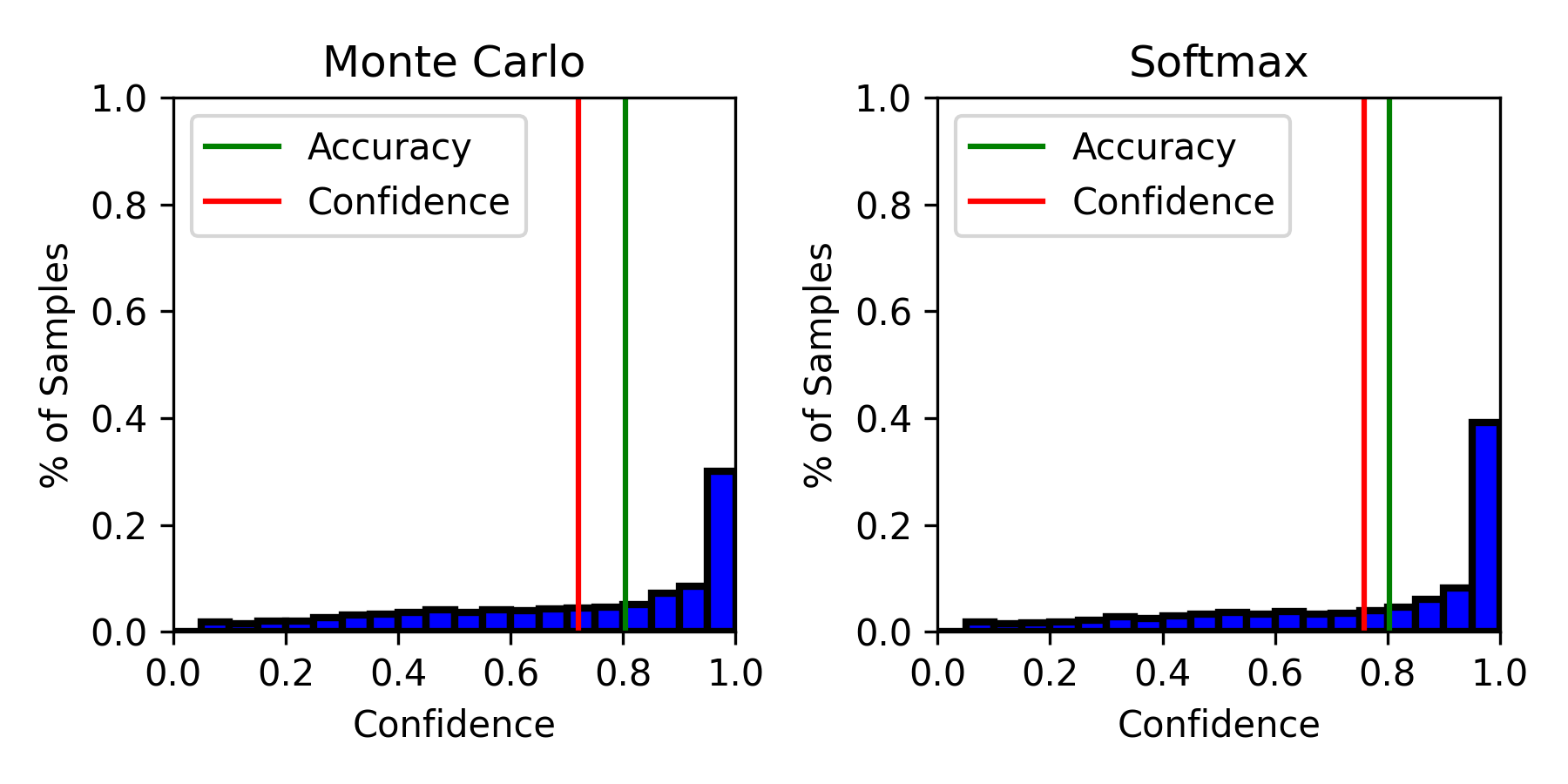}
    \end{minipage}
    \caption{Reliability diagram (left, displayed as a stacked bar chart comparing accuracy and confidence}) and confidence histogram (right) of 20 Newsgroups using GloVe-CNN. Comparing the plots of the figure to Figure \ref{fig:rel_conf_bert}, we see slight differences in both the reliability diagram and the confidence histogram. Most noticeable, we see slight differences in the reliability diagram, where we see more significant gaps between the confidence and the outputs, which indicates a less calibrated model due to the GloVe embeddings.
    \label{fig:rel_conf_glove}
\end{figure*}

\subsubsection{Performance Metrics}
\label{sssection:performance}
We use two main uncertainty metrics: test data holdout and expected calibration error (ECE). These metrics give us an estimation of the \textit{epistemic} uncertainty of the model i.e. the lack of certainty inherent in the model and its predictions. We do not cover metrics of \textit{aleatoric} uncertainty in this paper, which focus on the inherent randomness of the data itself and which could be tested through the introduction of e.g. label noise. For base model performance, we record the macro F1 score on the 20 Newsgroups, IMDb, Wiki and SST-2 datasets, and the accuracy on the Amazon dataset.

\noindent\textbf{Test data holdout:}
This metric ranks all samples based on the predictive uncertainty and calculates the F1 and accuracy scores on a percentage of the samples by removing those which the model is least certain about. In other words, a method is better if it achieves a greater improvement in performance metrics (e.g. F1) when removing the most uncertain samples. As such, this metric expresses the relationship between model calibration and accuracy. We choose to remove $10\%$, $20\%$, $30\%$ and $40\%$ of the least certain samples for our experiments. This metric shows how well the two methods can identify uncertain predictions of the model, as reflected by improvements in performance when more uncertain predictions are removed~\cite{zhang_mitigating_2019}. In our experiments, we use the former mentioned Mean MC, DE and softmax method to calculate the uncertainties; we further add the Penultimate Layer Variance (PL-Variance), where the PL-Variance utilises the variance of the last fully-connected layer as the uncertainty~\cite{zaragoza_confidence_1998}.

\noindent\textbf{Expected calibration error:}
As a second uncertainty estimation metric, we use the expected calibration error (ECE, \citet{guo_calibration_2017}), which measures in expectation, how confident are the predictions for both correct and incorrect predictions. This tells us how well each of the MC dropout and softmax methods estimate the uncertainties at the level of probability distributions, as opposed to the holdout method which only looks at downstream task performance.
ECE works by dividing the data into $m$ bins, where each bin in $B$ contains data that is within a certain range of probabilities, using the probability of the predicted class. Formally, ECE is defined as:
\begin{align}
    ECE &= \sum_{m=1}^{M} \frac{|B_m|}{n}|acc(B_m) - conf(B_m)|
\end{align}
where $M$ is the size of the dataset and $acc$ and $conf$ is the accuracy and mean confidence (i.e. predicted class probabilities) of the bin $B_m$.

Finally, to visualise the difference between the MC dropout and softmax, we create both confidence histograms and reliability diagrams~\cite{guo_calibration_2017}.
The reliability diagrams show how close the models are to perfect calibration, where perfect calibration means that the models accuracy and confidence is equal to the bins confidence range. In all cases, we show reliability diagrams by comparing histograms of accuracy and confidence across confidence bins; as such, when confidence exceeds accuracy in a given bin, that indicates how overconfident the model is for that bin.
The reliability diagrams help us visualise the ECE, by showing the accuracy and mean confidence of each bin, where each bin consists of the data which have a confidence within the range of the bin.
To complement the reliability diagrams, we also use confidence histograms, which show the distribution of confidence.

\subsection{Efficiency Results}
\label{sssection:efficiency}
In Table \ref{table:efficiency_holdout}, we display the runtime of the different model and method combinations. The runtime for the forward passes is calculated as a sum of all the forward passes on the entire dataset, and the runtime for the uncertainty methods are calculated for the entire dataset. Observing the results, we see that softmax is overall faster, and is approximately $10$ times faster when only looking at the forward passes, and using more complex aggregation methods in MC dropout, like DE, can be computationally heavy.

\begin{table}[h]
\centering
\scalebox{0.85}{
\begin{tabular}{@{}l@{\hspace{-1.0ex}}ccc@{}}
\toprule
 & Forward passes & Mean MC & DE \\ \midrule
20 Newsgroups & 1.0876 & 0.0003 & 12.3537 \\
IMDb & 1.386 & 0.0018 & 216.11 \\
Amazon & 4.9126 & 0.0017 & 194.08 \\
WIKI & 1.1149 & 0.0010 & 15.8467 \\
SST-2 & 1.0076 & 0.0003 & 3.4785 \\
\midrule
 & Forward passes & Softmax & PL-Variance \\ \midrule
20 Newsgroups & 0.0130 & 0.0002 & 0.0001 \\
IMDb & 0.0387 & 0.0003 & 0.0003 \\
Amazon & 0.4067 & 0.0004 & 0.0002 \\
WIKI & 0.0149 & 0.0002 & 0.0001 \\
SST-2 & 0.0037 & 0.0002 & 0.0001 \\ \bottomrule
\end{tabular}
}
\caption{Runtime measured in seconds for both MC dropout (top) and softmax (bottom). The times are on the full datasets split into the runtime of the forward passes and the runtime of calculating the uncertainty.}
\label{table:efficiency_holdout}
\end{table}

\begin{figure*}[t]
    \begin{minipage}[t]{0.49\linewidth}
        \centering
        \includegraphics[width=1.0\linewidth]{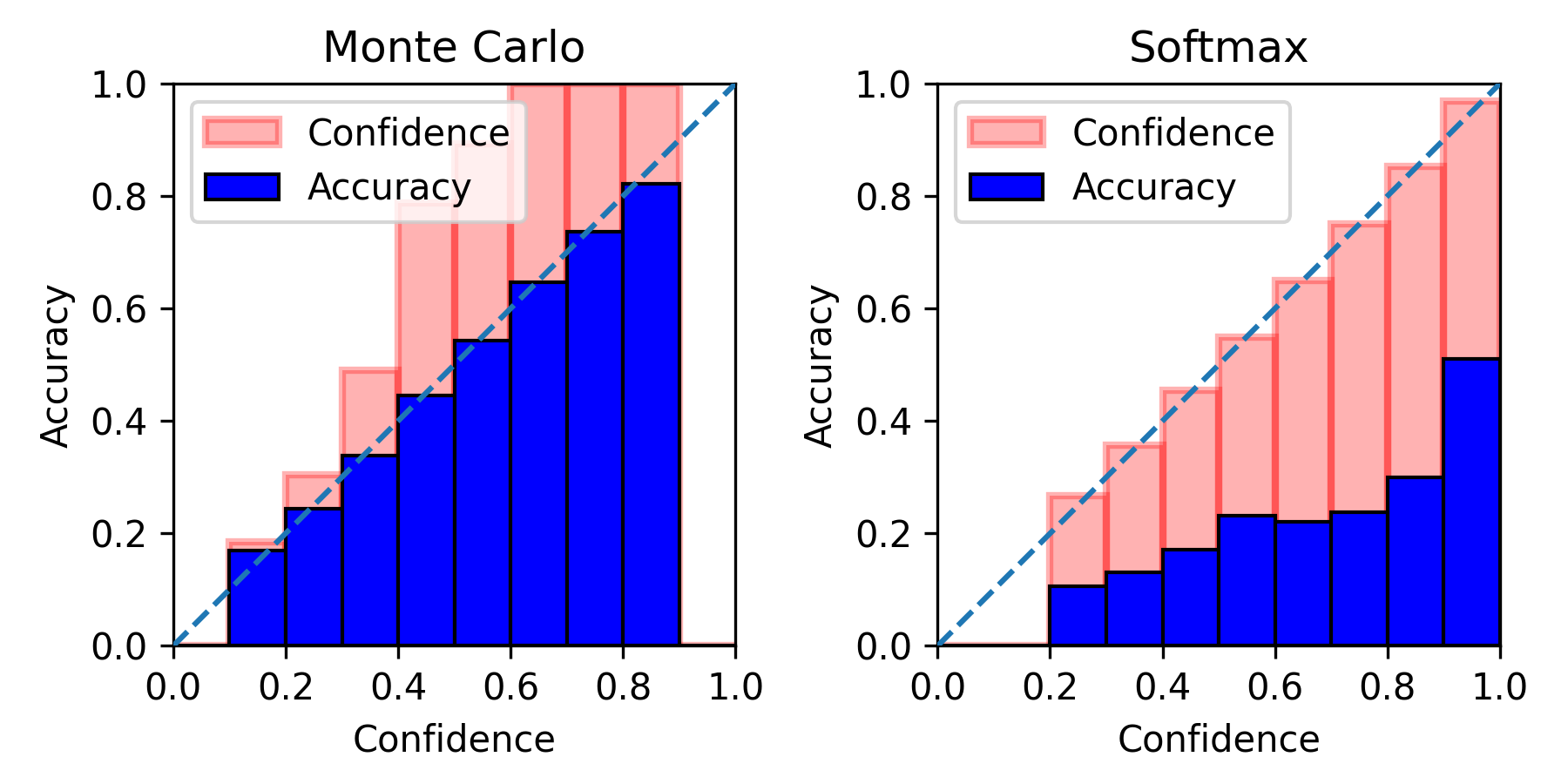}
    \end{minipage}
    \hfill
    \begin{minipage}[t]{0.49\linewidth}
        \centering
        \includegraphics[width=1.0\linewidth]{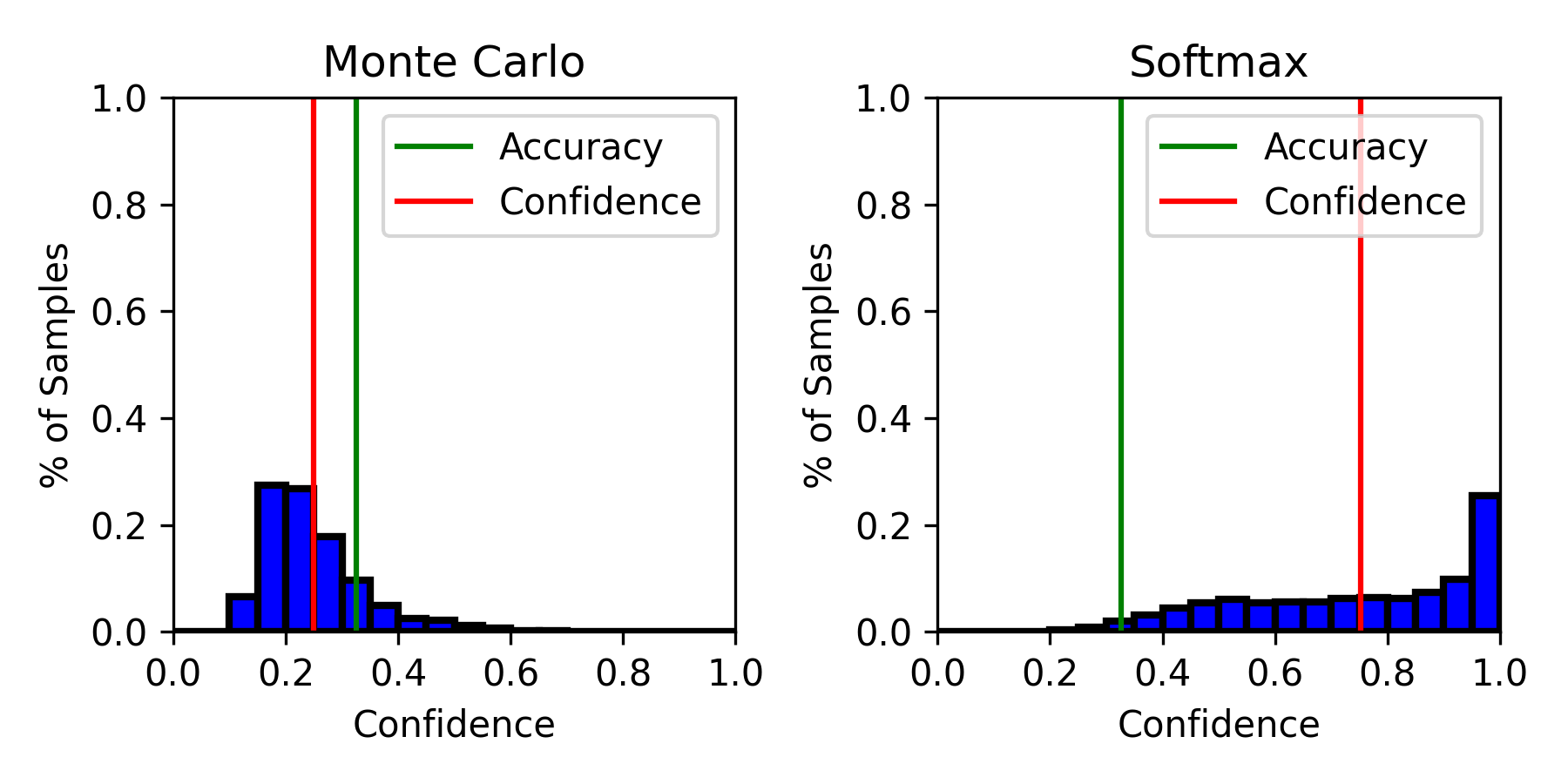}
    \end{minipage}
    \caption{Reliability diagram of 20 Newsgroups dataset (displayed as a stacked bar chart comparing accuracy and confidence) using the BERT-CNN model, with added zero-mean Gaussian noise to the BERT embeddings. Softmax is highly overconfident compared to MC dropout (despite the efficient setting in this paper where only the final layers of the model are used for dropout), as indicated by the large gap between average confidence and accuracy in each bin of the histogram.}
    \label{fig:rel_conf_bert_noise}
\end{figure*}

\subsection{Test Data Holdout Results}
Table~\ref{table:performance_holdout_20news} and the table in Appendix \ref{app:results} show the performance of the two uncertainty approximation methods using the different datasets and models. The tables show the macro F1 score and accuracy (depending on the datasets), and the ratio of improvement from holding out data in parentheses. We observe that in most cases, either dropout-entropy (DE) or softmax has the highest score and improvement ratio. However, in most cases the two are close in performance and improvement ratio. We further observe that Mean MC also performs well and is almost on par with DE, however, Mean MC is a much more efficient method compared to DE, so the slight trade-off in performance could be beneficial in resource-constrained settings or non-critical applications.

\begin{table*}[h]
\centering
\scalebox{0.84}{
\begin{tabular}{@{}lccccc@{}}
\toprule
\textbf{BERT} & 0\% & 10\% & 20\% & 30\% & 40\% \\ \midrule
Mean MC & 0.8591 & 0.8985 (1.0459) & 0.9225 (1.0739) & 0.9406 (1.0949) & 0.9487 (1.1043) \\
DE & 0.8591 & 0.9050 (1.0534) & 0.9390 (1.0930) & 0.9584 (1.1156) & 0.9703 (1.1294) \\
Softmax & \textbf{0.8576} & \textbf{0.9072} (\textbf{1.0578}) & \textbf{0.9452} (\textbf{1.1021}) & \textbf{0.9620} (\textbf{1.1216}) & \textbf{0.9742} (\textbf{1.1360}) \\
PL-Variance & \textbf{0.8576} & 0.9006 (1.0501) & 0.9246 (1.0781) & 0.9403 (1.0964) & 0.9484 (1.1058) \\
\midrule
\textbf{GloVe} \\
\midrule
Mean MC & \textbf{0.7966} & 0.8450 (1.0608) & 0.8674 (1.0888) & 0.8846 (0.1104) & 0.8960 (1.1248) \\
DE & \textbf{0.7966} & \textbf{0.8469} (1.0631) & \textbf{0.8855} (\textbf{1.1116}) & \textbf{0.9155} (1.1492) & \textbf{0.9416} (\textbf{1.1820}) \\
Softmax & 0.7959 & 0.8465 (\textbf{1.0636}) & 0.8846 (1.1115) & 0.9149 (\textbf{1.1496}) & 0.9402 (1.1813) \\
PL-Variance & 0.7959 & 0.8436 (1.0599) & 0.8667 (1.0891) & 0.8848 (1.1118) & 0.8966 (1.1266) \\ \bottomrule
\end{tabular}
} 
\caption{Macro F1 score and improvement rate for the 20 Newsgroups dataset.}
\label{table:performance_holdout_20news}
\end{table*}

\subsection{Model Calibration Results}
\label{ssection:model_calibration}
To further investigate the differences between MC dropout and softmax, we utilise the expected calibration error (ECE) to observe the differences in the predictive uncertainties. In Table \ref{table:ece_table}, we show the accuracy and ECE on the three datasets using the BERT embeddings.

\begin{table}[h]
\centering
\scalebox{0.85}{
\begin{tabular}{@{}lcc@{}}
\toprule
 & Accuracy & ECE \\ \midrule
20 Newsgroups - Mean MC & 0.8655 & 0.0275 \\
20 Newsgroups - Softmax & 0.8642 & 0.0253 \\
\midrule
IMDb - Mean MC & 0.9354 & 0.0061 \\
IMDb - Softmax & 0.9364 & 0.0043  \\
\midrule
Amazon - Mean MC & 0.7466 & 0.0083 \\
Amazon - Softmax & 0.7474 & 0.0097 \\
\midrule
WIKI - Mean MC & 0.9227 & 0.0370 \\
WIKI - Softmax & 0.9230 & 0.0279 \\
\midrule
SST-2 - Mean MC & 0.7408 & 0.0535 \\
SST-2 - Softmax & 0.7442 & 0.0472 \\
\bottomrule
\end{tabular}
}
\caption{Accuracy and ECE of the two uncertainty approximation approaches on the three selected datasets.}
\label{table:ece_table}
\end{table}

The results from our holdout experiments in Table \ref{table:performance_holdout_20news} and in Appendix \ref{app:results} combined with the results from our ECE calculations in Table \ref{table:ece_table}, all point in the direction of the efficient MC dropout used in this study and softmax performing on par to each other, but with a large gap in runtime as shown in Table \ref{table:efficiency_holdout}. To get a better understanding of if and where the two methods diverge, we plot the reliability diagrams and confidence histograms as described in Section \ref{sssection:performance}.

\noindent\textbf{Plot description:}
In Figures \ref{fig:rel_conf_bert} and \ref{fig:rel_conf_glove}, we show the reliability diagrams and the confidence histograms on the 20 Newsgroups dataset using both our BERT-CNN and GloVe-CNN with both the MC dropout method and softmax. We create the reliability diagrams using $10$ bins and the confidence histograms with $20$. Where the reliability diagram's and confidence histogram's bins are an interval of confidence. We use $20$ bins for the confidence histograms to obtain a more fine-grained view of the distribution. In the reliability diagram, the $x$-axis is the confidence and the $y$-axis is the accuracy. For the confidence histogram the $x$-axis is again the confidence and the $y$-axis is the percentage of the samples in the given bin.

\noindent\textbf{Expectations:}
While ECE can quantify the performance of the models on a somewhat lower level than our other metrics, the metric can be deceived, especially in cases where models score high in accuracy. It will favour overconfident models; therefore, we expect the results to favour softmax. Looking at the ECE, we can observe that it will favour an overconfident method when the model achieves high accuracy. With this in mind, we expect the results to be skewed towards the softmax.

\noindent\textbf{Observations reliability diagram:}
From the reliability diagram, we observe that the difference in confidence and outputs are small. The difference between the two uncertainty methods is also minimal, including both BERT and GloVe embeddings, suggesting minimal potential gains from using MC dropout in an efficient setting while still incurring a high cost in terms of runtime. We determine that there is minimal difference by visually inspecting the plots, and by observing the ECE displayed in Table \ref{table:ece_table}. We further observe that in both MC dropout and softmax that the model worsens when we use the GloVe embeddings.

\noindent\textbf{Observations confidence histogram:}
As mentioned earlier, we know that the softmax tends to be overconfident, which can be seen in the percentage of samples in the last bin. The MC dropout method, on the other hand, utilises the probability space to a greater extent.
We include reliability diagrams and confidence histograms for the $2$ other datasets in Appendix \ref{app:model_calibration}.

\noindent\textbf{Noise experiment:}
Inspecting both Table \ref{table:ece_table} showing the ECE values and the performances in Table \ref{table:performance_holdout_20news}, \ref{table:performance_holdout_imdb} and \ref{table:performance_holdout_amazon}, we observe that using our two uncertainty estimation methods we achieved very high F1 scores and accuracies and low ECEs. We hypothesised that high performance could lead to softmax achieving high ECE, due to naturally having high confidence, compared to MC dropout. We added zero-mean Gaussian noise to the 20 Newsgroups test embeddings and redid our ECE experiments to test our hypothesis. In Figure \ref{fig:rel_conf_bert_noise}, we show the reliability diagram of the experiment with added noise, which shows the MC dropout outperforming softmax. To further build on the theory, we also inspect the confidence histogram, showing that softmax is still overconfident and the difference between the accuracy and mean confidence is high. This suggests that MC dropout is more resilient to noise, and in cases where the performance of a model is low, MC dropout could potentially obtain more precise predictive uncertainties.

\section{Discussion and Conclusion}
In this paper, we perform an in-depth empirical comparison of using the MC dropout method in an efficient setting and the more straightforward softmax method.
By doing a thorough empirical analysis of the two methods, shown in Section \ref{sssection:performance}, using various metrics to measure their performance on both efficiency and performance levels, we see that in our holdout experiments in table \ref{table:performance_holdout_20news}, that the two methods perform approximately the same.

Looking at the expected calibration error (ECE) experiments, the results again show that the MC dropout and softmax method perform somewhat equally, which we have shown in Section \ref{ssection:model_calibration}. We observe differences in the results as we observe a lower accuracy score, which we show in our noise experiment, which is also shown in Section \ref{ssection:model_calibration}.

Prior research \cite{hendrycks_baseline_2018} investigated out-of-distribution analysis and found that softmax, both for sentiment classification and text categorisation tasks, can detect out-of-distribution data points efficiently. It further showcases that in these two tasks, that the softmax also to some extent, can perform well as a confidence estimator.

While we show that the two methods perform almost equally, when comparing the predictive performance, the cost of using MC dropout is at a minimum $10$ times that of running softmax even in the efficient setting where only the final layer is dropped out, depending on the post-processing of the uncertainties, as we show in Section \ref{sssection:efficiency}. The post-processing cost of MC dropout can quickly explode when used on larger datasets or if a more expensive method like dropout-entropy is used instead of simpler approaches.

Given this, when could it be appropriate to use the more efficient softmax over MC dropout for estimating predictive uncertainty? Our results suggest that when the base accuracy of a model is high, the differences in uncertainty estimation between the two methods is relatively low, likely due to the higher confidence of the softmax method. In this case, if latency or resource efficiency is a concern such as on edge devices, it may be appropriate to rely on a quick estimate using softmax as opposed to a more cumbersome method. However, when model accuracy is expected to be low, softmax is still overconfident compared to MC dropout, so estimates using a single deterministic softmax may be unreliable. The downstream application may also impact this: in critical scenarios such as health care, it may still be more appropriate to use an inefficient method with better predictive uncertainty for improved decision-making. In low-risk applications where models are known to be accurate and efficiency is of concern, we have demonstrated that softmax can potentially be sufficient.

\section*{Limitations} We highlight a few key limitations of the study to further contextualise the work. First, we note that the study is restricted to neural network based methods, while other methods in ML may be useful to study for uncertainty estimation as well. Second, we note that we test a plain softmax method without temperature scaling -- while calibrating a useful temperature could induce a cost in terms of time, it would potentially lead to better uncertainty estimation. Finally, we note that we also test an efficient form of MC dropout which only drops out a portion of the network. While this demonstrates that in an efficient setting, softmax can be as good or better at uncertainty estimation than MC dropout, full MC dropout still may have better uncertainty estimation when efficiency is not a concern.

\section*{Acknowledgements}
This research was funded by Innovation Fund Denmark grant number 9065-00131B.

\bibliographystyle{acl_natbib}
\bibliography{anthology,custom}

\begin{thebibliography}{34}
\expandafter\ifx\csname natexlab\endcsname\relax\def\natexlab#1{#1}\fi

\bibitem[{Blundell et~al.(2015)Blundell, Cornebise, Kavukcuoglu, and
  Wierstra}]{blundell_weight_2015}
Charles Blundell, Julien Cornebise, Koray Kavukcuoglu, and Daan Wierstra. 2015.
\newblock \href {https://proceedings.mlr.press/v37/blundell15.html} {Weight
  {Uncertainty} in {Neural} {Network}}.
\newblock In \emph{Proceedings of the 32nd {International} {Conference} on
  {Machine} {Learning}}, pages 1613--1622. PMLR.
\newblock ISSN: 1938-7228.

\bibitem[{Devlin et~al.(2019)Devlin, Chang, Lee, and
  Toutanova}]{devlin_bert_2019}
Jacob Devlin, Ming-Wei Chang, Kenton Lee, and Kristina Toutanova. 2019.
\newblock \href {http://arxiv.org/abs/1810.04805} {{BERT}: {Pre}-training of
  {Deep} {Bidirectional} {Transformers} for {Language} {Understanding}}.
\newblock \emph{arXiv:1810.04805 [cs]}.
\newblock ArXiv: 1810.04805.

\bibitem[{Durasov et~al.(2021)Durasov, Bagautdinov, Baque, and
  Fua}]{durasov2021masksembles}
Nikita Durasov, Timur Bagautdinov, Pierre Baque, and Pascal Fua. 2021.
\newblock Masksembles for uncertainty estimation.
\newblock In \emph{Proceedings of the IEEE/CVF Conference on Computer Vision
  and Pattern Recognition}, pages 13539--13548.

\bibitem[{Gal and Ghahramani(2016{\natexlab{a}})}]{gal_dropout_app_2016}
Yarin Gal and Zoubin Ghahramani. 2016{\natexlab{a}}.
\newblock \href {http://arxiv.org/abs/1506.02157} {Dropout as a {Bayesian}
  {Approximation}: {Appendix}}.
\newblock \emph{arXiv:1506.02157 [stat]}.
\newblock ArXiv: 1506.02157.

\bibitem[{Gal and Ghahramani(2016{\natexlab{b}})}]{gal_dropout_2016}
Yarin Gal and Zoubin Ghahramani. 2016{\natexlab{b}}.
\newblock \href {https://proceedings.mlr.press/v48/gal16.html} {Dropout as a
  {Bayesian} {Approximation}: {Representing} {Model} {Uncertainty} in {Deep}
  {Learning}}.
\newblock In \emph{Proceedings of {The} 33rd {International} {Conference} on
  {Machine} {Learning}}, pages 1050--1059. PMLR.
\newblock ISSN: 1938-7228.

\bibitem[{Gawlikowski et~al.(2021)Gawlikowski, Tassi, Ali, Lee, Humt, Feng,
  Kruspe, Triebel, Jung, Roscher, Shahzad, Yang, Bamler, and
  Zhu}]{gawlikowski_survey_2021}
Jakob Gawlikowski, Cedrique Rovile~Njieutcheu Tassi, Mohsin Ali, Jongseok Lee,
  Matthias Humt, Jianxiang Feng, Anna Kruspe, Rudolph Triebel, Peter Jung,
  Ribana Roscher, Muhammad Shahzad, Wen Yang, Richard Bamler, and Xiao~Xiang
  Zhu. 2021.
\newblock \href {http://arxiv.org/abs/2107.03342} {A {Survey} of {Uncertainty}
  in {Deep} {Neural} {Networks}}.
\newblock \emph{arXiv:2107.03342 [cs, stat]}.
\newblock ArXiv: 2107.03342.

\bibitem[{Guo et~al.(2017)Guo, Pleiss, Sun, and
  Weinberger}]{guo_calibration_2017}
Chuan Guo, Geoff Pleiss, Yu~Sun, and Kilian~Q. Weinberger. 2017.
\newblock \href {https://proceedings.mlr.press/v70/guo17a.html} {On
  {Calibration} of {Modern} {Neural} {Networks}}.
\newblock In \emph{Proceedings of the 34th {International} {Conference} on
  {Machine} {Learning}}, pages 1321--1330. PMLR.
\newblock ISSN: 2640-3498.

\bibitem[{He et~al.(2020)He, Zhang, Lei, Chen, Chen, Alhamadani, Xiao, and
  Lu}]{he-etal-2020-towards}
Jianfeng He, Xuchao Zhang, Shuo Lei, Zhiqian Chen, Fanglan Chen, Abdulaziz
  Alhamadani, Bei Xiao, and ChangTien Lu. 2020.
\newblock \href {https://doi.org/10.18653/v1/2020.emnlp-main.671} {Towards more
  accurate uncertainty estimation in text classification}.
\newblock In \emph{Proceedings of the 2020 Conference on Empirical Methods in
  Natural Language Processing (EMNLP)}, pages 8362--8372, Online. Association
  for Computational Linguistics.

\bibitem[{Hendrycks and Gimpel(2017)}]{hendrycks_baseline_2018}
Dan Hendrycks and Kevin Gimpel. 2017.
\newblock \href {https://openreview.net/forum?id=Hkg4TI9xl} {A baseline for
  detecting misclassified and out-of-distribution examples in neural networks}.
\newblock In \emph{5th International Conference on Learning Representations,
  {ICLR} 2017, Toulon, France, April 24-26, 2017, Conference Track
  Proceedings}. OpenReview.net.

\bibitem[{Henne et~al.(2020)Henne, Schwaiger, Roscher, and
  Weiss}]{conf/aaai/HenneSRW20}
Maximilian Henne, Adrian Schwaiger, Karsten Roscher, and Gereon Weiss. 2020.
\newblock \href {http://ceur-ws.org/Vol-2560/paper35.pdf} {Benchmarking
  uncertainty estimation methods for deep learning with safety-related
  metrics}.
\newblock In \emph{Proceedings of the Workshop on Artificial Intelligence
  Safety, co-located with 34th {AAAI} Conference on Artificial Intelligence,
  SafeAI@AAAI 2020, New York City, NY, USA, February 7, 2020}, volume 2560 of
  \emph{{CEUR} Workshop Proceedings}, pages 83--90. CEUR-WS.org.

\bibitem[{Hinton and van Camp(1993)}]{hinton_keeping_1993}
Geoffrey~E. Hinton and Drew van Camp. 1993.
\newblock \href {https://doi.org/10.1145/168304.168306} {Keeping the neural
  networks simple by minimizing the description length of the weights}.
\newblock In \emph{Proceedings of the sixth annual conference on
  {Computational} learning theory}, {COLT} '93, pages 5--13, New York, NY, USA.
  Association for Computing Machinery.

\bibitem[{Joo et~al.(2020)Joo, Chung, and Seo}]{joo_being_2020}
Taejong Joo, Uijung Chung, and Min-Gwan Seo. 2020.
\newblock \href {https://proceedings.mlr.press/v119/joo20a.html} {Being
  {Bayesian} about {Categorical} {Probability}}.
\newblock In \emph{Proceedings of the 37th {International} {Conference} on
  {Machine} {Learning}}, pages 4950--4961. PMLR.
\newblock ISSN: 2640-3498.

\bibitem[{Kingma and Ba(2015)}]{kingma_adam_2015}
Diederik~P. Kingma and Jimmy Ba. 2015.
\newblock \href {http://arxiv.org/abs/1412.6980} {Adam: {A} method for
  stochastic optimization}.
\newblock In \emph{3rd International Conference on Learning Representations,
  {ICLR} 2015, San Diego, CA, USA, May 7-9, 2015, Conference Track
  Proceedings}.

\bibitem[{Lakshminarayanan et~al.(2017)Lakshminarayanan, Pritzel, and
  Blundell}]{conf/nips/Lakshminarayanan17}
Balaji Lakshminarayanan, Alexander Pritzel, and Charles Blundell. 2017.
\newblock \href
  {https://proceedings.neurips.cc/paper/2017/file/9ef2ed4b7fd2c810847ffa5fa85bce38-Paper.pdf}
  {Simple and scalable predictive uncertainty estimation using deep ensembles}.
\newblock In \emph{Advances in Neural Information Processing Systems},
  volume~30. Curran Associates, Inc.

\bibitem[{Lang(1995)}]{lang_newsweeder_1995}
Ken Lang. 1995.
\newblock \href {https://doi.org/10.1016/B978-1-55860-377-6.50048-7}
  {{NewsWeeder}: {Learning} to {Filter} {Netnews}}.
\newblock In Armand Prieditis and Stuart Russell, editors, \emph{Machine
  {Learning} {Proceedings} 1995}, pages 331--339. Morgan Kaufmann, San
  Francisco (CA).

\bibitem[{Lin et~al.(2021)Lin, Wang, Liu, and Qiu}]{lin2021survey}
Tianyang Lin, Yuxin Wang, Xiangyang Liu, and Xipeng Qiu. 2021.
\newblock \href {http://arxiv.org/abs/2106.04554} {A survey of transformers}.
\newblock \emph{CoRR}, abs/2106.04554.

\bibitem[{Maas et~al.(2011)Maas, Daly, Pham, Huang, Ng, and
  Potts}]{maas_learning_2011}
Andrew~L. Maas, Raymond~E. Daly, Peter~T. Pham, Dan Huang, Andrew~Y. Ng, and
  Christopher Potts. 2011.
\newblock \href {https://www.aclweb.org/anthology/P11-1015} {Learning {Word}
  {Vectors} for {Sentiment} {Analysis}}.
\newblock In \emph{Proceedings of the 49th {Annual} {Meeting} of the
  {Association} for {Computational} {Linguistics}: {Human} {Language}
  {Technologies}}, pages 142--150, Portland, Oregon, USA. Association for
  Computational Linguistics.

\bibitem[{MacKay(1992)}]{mackay_practical_1992}
David J.~C. MacKay. 1992.
\newblock \href {https://doi.org/10.1162/neco.1992.4.3.448} {A {Practical}
  {Bayesian} {Framework} for {Backpropagation} {Networks}}.
\newblock \emph{Neural Computation}, 4(3):448--472.

\bibitem[{McAuley and Leskovec(2013)}]{mcauley_hidden_2013}
Julian McAuley and Jure Leskovec. 2013.
\newblock \href {https://doi.org/10.1145/2507157.2507163} {Hidden factors and
  hidden topics: understanding rating dimensions with review text}.
\newblock In \emph{Proceedings of the 7th {ACM} conference on {Recommender}
  systems}, {RecSys} '13, pages 165--172, New York, NY, USA. Association for
  Computing Machinery.

\bibitem[{Mozejko et~al.(2018)Mozejko, Susik, and
  Karczewski}]{mozejko2019inhibited}
Marcin Mozejko, Mateusz Susik, and Rafal Karczewski. 2018.
\newblock \href {http://arxiv.org/abs/1810.01861} {Inhibited softmax for
  uncertainty estimation in neural networks}.
\newblock \emph{CoRR}, abs/1810.01861.

\bibitem[{Naeini et~al.(2015)Naeini, Cooper, and
  Hauskrecht}]{naeini_obtaining_2015}
Mahdi~Pakdaman Naeini, Gregory~F. Cooper, and Milos Hauskrecht. 2015.
\newblock Obtaining well calibrated probabilities using bayesian binning.
\newblock 2015:2901--2907.

\bibitem[{Neal(1993)}]{neal_bayesian_1993}
Radford Neal. 1993.
\newblock Bayesian {Training} of {Backpropagation} {Networks} by the {Hybrid}
  {Monte} {Carlo} {Method}.
\newblock Technical report.

\bibitem[{Niculescu-Mizil and Caruana(2015)}]{niculescu-mizil_predicting_2005}
Alexandru Niculescu-Mizil and Rich Caruana. 2015.
\newblock \href {https://doi.org/10.1145/1102351.1102430} {Predicting good
  probabilities with supervised learning}.
\newblock In \emph{Proceedings of the 22nd international conference on Machine
  learning}, {ICML} '05, pages 625--632. Association for Computing Machinery.

\bibitem[{Ovadia et~al.(2019)Ovadia, Fertig, Ren, Nado, Sculley, Nowozin,
  Dillon, Lakshminarayanan, and Snoek}]{ovadia_can_2019}
Yaniv Ovadia, Emily Fertig, Jie Ren, Zachary Nado, D.~Sculley, Sebastian
  Nowozin, Joshua Dillon, Balaji Lakshminarayanan, and Jasper Snoek. 2019.
\newblock \href
  {https://proceedings.neurips.cc/paper/2019/hash/8558cb408c1d76621371888657d2eb1d-Abstract.html}
  {Can you trust your model' s uncertainty? {Evaluating} predictive uncertainty
  under dataset shift}.
\newblock In \emph{Advances in {Neural} {Information} {Processing} {Systems}},
  volume~32. Curran Associates, Inc.

\bibitem[{Patterson et~al.(2021)Patterson, Gonzalez, Le, Liang, Munguia,
  Rothchild, So, Texier, and Dean}]{patterson2021carbon}
David~A. Patterson, Joseph Gonzalez, Quoc~V. Le, Chen Liang, Lluis{-}Miquel
  Munguia, Daniel Rothchild, David~R. So, Maud Texier, and Jeff Dean. 2021.
\newblock \href {http://arxiv.org/abs/2104.10350} {Carbon emissions and large
  neural network training}.
\newblock \emph{CoRR}, abs/2104.10350.

\bibitem[{Pennington et~al.(2014)Pennington, Socher, and
  Manning}]{pennington_glove_2014}
Jeffrey Pennington, Richard Socher, and Christopher~D. Manning. 2014.
\newblock \href {https://doi.org/10.3115/v1/d14-1162} {Glove: Global vectors
  for word representation}.
\newblock In \emph{Proceedings of the 2014 Conference on Empirical Methods in
  Natural Language Processing, {EMNLP} 2014, October 25-29, 2014, Doha, Qatar,
  {A} meeting of SIGDAT, a Special Interest Group of the {ACL}}, pages
  1532--1543. {ACL}.

\bibitem[{Redi et~al.(2019)Redi, Fetahu, Morgan, and
  Taraborelli}]{redi_citation_2019}
Miriam Redi, Besnik Fetahu, Jonathan~T. Morgan, and Dario Taraborelli. 2019.
\newblock \href {http://arxiv.org/abs/1902.11116} {Citation needed: {A}
  taxonomy and algorithmic assessment of wikipedia's verifiability}.
\newblock \emph{CoRR}, abs/1902.11116.

\bibitem[{Socher et~al.(2013)Socher, Perelygin, Wu, Chuang, Manning, Ng, and
  Potts}]{socher_recursive_2013}
Richard Socher, Alex Perelygin, Jean Wu, Jason Chuang, Christopher~D. Manning,
  Andrew~Y. Ng, and Christopher Potts. 2013.
\newblock \href {https://aclanthology.org/D13-1170/} {Recursive deep models for
  semantic compositionality over a sentiment treebank}.
\newblock In \emph{Proceedings of the 2013 Conference on Empirical Methods in
  Natural Language Processing, {EMNLP} 2013, 18-21 October 2013, Grand Hyatt
  Seattle, Seattle, Washington, USA, {A} meeting of SIGDAT, a Special Interest
  Group of the {ACL}}, pages 1631--1642. {ACL}.

\bibitem[{Strubell et~al.(2019)Strubell, Ganesh, and
  McCallum}]{strubell-etal-2019-energy}
Emma Strubell, Ananya Ganesh, and Andrew McCallum. 2019.
\newblock \href {https://doi.org/10.18653/v1/P19-1355} {Energy and policy
  considerations for deep learning in {NLP}}.
\newblock In \emph{Proceedings of the 57th Annual Meeting of the Association
  for Computational Linguistics}, pages 3645--3650, Florence, Italy.
  Association for Computational Linguistics.

\bibitem[{Tay et~al.(2021)Tay, Dehghani, Gupta, Aribandi, Bahri, Qin, and
  Metzler}]{tay-etal-2021-pretrained}
Yi~Tay, Mostafa Dehghani, Jai~Prakash Gupta, Vamsi Aribandi, Dara Bahri, Zhen
  Qin, and Donald Metzler. 2021.
\newblock \href {https://doi.org/10.18653/v1/2021.acl-long.335} {Are pretrained
  convolutions better than pretrained transformers?}
\newblock In \emph{Proceedings of the 59th Annual Meeting of the Association
  for Computational Linguistics and the 11th International Joint Conference on
  Natural Language Processing (Volume 1: Long Papers)}, pages 4349--4359,
  Online. Association for Computational Linguistics.

\bibitem[{van Amersfoort et~al.(2020)van Amersfoort, Smith, Teh, and
  Gal}]{vanamersfoort2020uncertainty}
Joost van Amersfoort, Lewis Smith, Yee~Whye Teh, and Yarin Gal. 2020.
\newblock \href {http://proceedings.mlr.press/v119/van-amersfoort20a.html}
  {Uncertainty estimation using a single deep deterministic neural network}.
\newblock In \emph{Proceedings of the 37th International Conference on Machine
  Learning, {ICML} 2020, 13-18 July 2020, Virtual Event}, volume 119 of
  \emph{Proceedings of Machine Learning Research}, pages 9690--9700. {PMLR}.

\bibitem[{Vaswani et~al.(2017)Vaswani, Shazeer, Parmar, Uszkoreit, Jones,
  Gomez, Kaiser, and Polosukhin}]{vaswani2017attention}
Ashish Vaswani, Noam Shazeer, Niki Parmar, Jakob Uszkoreit, Llion Jones,
  Aidan~N. Gomez, Lukasz Kaiser, and Illia Polosukhin. 2017.
\newblock \href {https://papers.nips.cc/paper/7181-attention-is-all-you-need}
  {Attention is all you need}.
\newblock In I.~Guyon, U.~V. Luxburg, S.~Bengio, H.~Wallach, R.~Fergus,
  S.~Vishwanathan, and R.~Garnett, editors, \emph{Advances in Neural
  Information Processing Systems 30}, page 5998–6008. Curran Associates, Inc.

\bibitem[{Zaragoza and d'Alch{\'e} Buc(1998)}]{zaragoza_confidence_1998}
Hugo Zaragoza and Florence d'Alch{\'e} Buc. 1998.
\newblock \href {https://hal.science/hal-01622612} {{Confidence Measures for
  Neural Network Classifiers}}.
\newblock In \emph{{7th Conference on Information Processing and Management of
  Uncertainty in Knowledge-Based Systems}}, Paris, France.

\bibitem[{Zhang et~al.(2019)Zhang, Chen, Lu, and
  Ramakrishnan}]{zhang_mitigating_2019}
Xuchao Zhang, Fanglan Chen, Chang-Tien Lu, and Naren Ramakrishnan. 2019.
\newblock \href {https://doi.org/10.18653/v1/N19-1316} {Mitigating
  {Uncertainty} in {Document} {Classification}}.
\newblock In \emph{Proceedings of the 2019 {Conference} of the {North}
  {American} {Chapter} of the {Association} for {Computational} {Linguistics}:
  {Human} {Language} {Technologies}, {Volume} 1 ({Long} and {Short} {Papers})},
  pages 3126--3136, Minneapolis, Minnesota. Association for Computational
  Linguistics.

\end{thebibliography}

\newpage

\appendix
\section{Reproducibility}

\subsection{Computing Infrastructure}
All Experiments were run on a Microsoft Azure NC6-series server. With the following specifications: 6 Inter Xeon-E5-2690 v3, NVIDIA Tesla K80 with 12GB RAM and 56GB of RAM.

\subsection{Hyperparameters}
We used the following hyperparameters for training our CNN model and CNN GloVe model: Epochs: 1000; batch size: 256 for 20 Newsgroups, IMDb SST-2 and Wiki and 128 for Amazon; early stopping: 10; learning rate: 0.001. For fine-tuning BERT we used the following set of hyperparamaters: epochs: 3; warm-up steps 500; weight decay 0.01; batch size 8; masked language model probability: 0.15. All hyperparameters are set without performing cross-validation.

\subsection{Dropout - Hyperparameter}
\label{app:dropout}
The performance of the MC dropout method is correlated with the dropout probability. We therefore run our CNN model using BERT embeddings on the 20 Newsgroups dataset with the following dropout probabilities $[0.1, 0.2, 0.3, 0.4, 0.5]$. In Table \ref{table:dropouts}, we show the results using the 5 different dropout probabilities, where we see that it stops improving at $0.4$ and $0.5$ percentage dropout. As such, we use a dropout of $0.5$ for our experiments.
\begin{table}[!h]
\centering
\scalebox{0.85}{
\begin{tabular}{@{}lccccc@{}}
\toprule
 & 0\% & 10\% & 20\% & 30\% & 40\% \\ \midrule
0.1 & 0.8598 & 0.9010 & 0.9255 & 0.9408 & 0.9483 \\
0.2 & 0.8599 & 0.9005 & 0.9256 & 0.9408 & 0.9502 \\
0.3 & 0.8596 & 0.9007 & 0.9245 & 0.9412 & 0.9491 \\
0.4 & 0.8601 & 0.8996 & 0.9253 & 0.9425 & 0.9502 \\
0.5 & 0.8591 & 0.8985 & 0.9225 & 0.9406 & 0.9487 \\ \bottomrule
\end{tabular}
}
\caption{We test how the dropout probabilities correlate with the performance of MC dropout, using the BERT-CNN model. The results are reported in terms of macro F1.}
\label{table:dropouts}
\end{table}

\section{Result tables}
\label{app:results}

\begin{table*}[!h]
\centering
\scalebox{0.84}{
\begin{tabular}{@{}lccccc@{}}
\toprule
\textbf{BERT} & 0\% & 10\% & 20\% & 30\% & 40\% \\ \midrule
Mean MC & 0.9354 & 0.9668 (1.0335) & 0.9829 (1.0508) & 0.9901 (1.0585) & 0.9930 (1.0616) \\
DE & 0.9354 & 0.9679 (1.0347) & 0.9789 (1.0465) & 0.9787 (1.0463) & 0.9798 (1.0475) \\
Softmax & \textbf{0.9364} & \textbf{0.9691} (\textbf{1.0349}) & \textbf{0.9847} (\textbf{1.0516}) & \textbf{0.9913} (\textbf{1.0586}) & \textbf{0.9940} (1.0615) \\
PL-Variance & \textbf{0.9364} & 0.9678 (1.0335) & 0.9837 (1.0506) & 0.9901 (1.0574) & 0.9933 (1.0608) \\
\midrule
\textbf{GloVe} \\
\midrule
Mean MC & \textbf{0.8825} & 0.9170 (1.0391) & 0.9416 (1.0670) & \textbf{0.9614} (\textbf{1.0894}) & 0.9730 (1.1025) \\
DE & \textbf{0.8825} & \textbf{0.9183} (\textbf{1.0406}) & \textbf{0.9430} (\textbf{1.0686}) & 0.9449 (1.0707) & 0.9455 (1.0714) \\
Softmax & 0.8824 & 0.9154 (1.0374) & 0.9406 (1.0660) & 0.9598 (1.0878) & 0.9724 (1.1020) \\
PL-Variance & 0.8824 & 0.9162 (1.0383) & 0.9415 (1.0670) & 0.9611 (1.0892) & \textbf{0.9736} (\textbf{1.1034}) \\ \bottomrule
\end{tabular}
} 
\caption{Macro F1 score and improvement rate for the IMDb dataset.}
\label{table:performance_holdout_imdb}
\end{table*}

\begin{table*}[!h]
\centering
\scalebox{0.84}{
\begin{tabular}{@{}lccccc@{}}
\toprule
\textbf{BERT} & 0\% & 10\% & 20\% & 30\% & 40\% \\ \midrule
Mean MC & 0.7466 & 0.7853 (1.0518) & 0.8137 (1.0898) & 0.8392 (1.1240) & 0.8605 (1.1526) \\
DE & 0.7466 & 0.7850 (1.0513) & 0.8191 (1.0871) & 0.8492 (1.1374) & 0.8684 (1.1631) \\
Softmax & \textbf{0.7474} & \textbf{0.7875} (\textbf{1.0537}) & \textbf{0.8225} (\textbf{1.1005}) & \textbf{0.8562} (\textbf{1.1456}) & \textbf{0.8845} (\textbf{1.1834}) \\
PL-Variance & \textbf{0.7474} & 0.7856 (1.0510) & 0.8144 (1.0896) & 0.8404 (1.1244) & 0.8610 (1.1520) \\
\midrule
\textbf{GloVe} \\
\midrule
Mean MC & 0.6979 & 0.7369 (\textbf{1.0559}) & 0.7675 (1.0998) & 0.7962 (1.1408) & 0.8214 (1.1770) \\
DE & 0.6979 & 0.7366 (1.0555) & 0.7716 (1.1056) & 0.8019 (1.1490) & 0.8102 (1.1610) \\
Softmax & \textbf{0.6984} & \textbf{0.7374} (\textbf{1.0559}) & \textbf{0.7730} (\textbf{1.1068}) & \textbf{0.8067} (\textbf{1.1550}) & \textbf{0.8359} (\textbf{1.1969}) \\
PL-Variance & \textbf{0.6984} & 0.7358 (1.0536) & 0.7676 (1.0990) & 0.7961 (1.1398) & 0.8209 (1.1753) \\ \bottomrule
\end{tabular}
} 
\caption{Accuracy score and improvement rate for the Amazon (Sports and Outdoors) dataset.}
\label{table:performance_holdout_amazon}
\end{table*}

\begin{table*}[!h]
\centering
\scalebox{0.84}{
\begin{tabular}{@{}lccccc@{}}
\toprule
\textbf{BERT} & 0\% & 10\% & 20\% & 30\% & 40\% \\ \midrule
Mean MC & 0.9227 & \textbf{0.9569} (\textbf{1.0370}) & 0.9742 (1.0557) & 0.9824 (1.0646) & \textbf{0.9878} (\textbf{1.0705}) \\
DE & 0.9227 & 0.9566 (1.0367) & 0.9743 (1.0559) & 0.9767 (1.0585) & 0.9762 (1.0579) \\
Softmax & \textbf{0.9230} & 0.9561 (1.0358) & 0.9745 (1.0558) & \textbf{0.9834} (\textbf{1.0655}) & 0.9869 (1.0692) \\
PL-Variance & \textbf{0.9230} & 0.9566 (1.0364) & \textbf{0.9748} (\textbf{1.0561}) & 0.9827 (1.0647) & 0.9869 (1.0693) \\
\midrule
\textbf{GloVe} \\
\midrule
Mean MC & \textbf{0.8559} & \textbf{0.8958} (1.0466) & 0.9168 (1.0712) & \textbf{0.9325} (1.0896) & 0.9379 (1.0958) \\
DE & \textbf{0.8559} & 0.8914 (1.0415) & 0.9146 (1.0686) & 0.9269 (1.0830) & 0.9319 (1.0889) \\
Softmax & 0.8539 & 0.8941 (1.0471) & 0.9181 (1.0752) & 0.9312 (1.0906) & \textbf{0.9393} (\textbf{1.1001}) \\
PL-Variance & 0.8539 & \textbf{0.8958} (\textbf{1.0491}) & \textbf{0.9209} (\textbf{1.0785}) & 0.9322 (\textbf{1.0918}) & 0.9366 (1.0969) \\ \bottomrule
\end{tabular}
} 
\caption{Macro F1 score and improvement rate for the Wiki dataset.}
\label{table:performance_holdout_wiki}
\end{table*}

\begin{table*}[!h]
\centering
\scalebox{0.84}{
\begin{tabular}{@{}lccccc@{}}
\toprule
\textbf{BERT} & 0\% & 10\% & 20\% & 30\% & 40\% \\ \midrule
Mean MC & 0.7407 & 0.7706 (1.0403) & 0.7907 (1.0674) & 0.8149 (1.1001) & 0.8432 (1.1383) \\
DE & 0.7407 & \textbf{0.7744} (\textbf{1.0454}) & \textbf{0.8008} (\textbf{1.0811}) & \textbf{0.8265} (\textbf{1.1158}) & \textbf{0.8472} (\textbf{1.1437}) \\
Softmax & \textbf{0.7442} & 0.7706 (1.0354) & 0.8006 (1.0758) & 0.8246 (1.1080) & 0.8451 (1.1355) \\
PL-Variance & \textbf{0.7442} & 0.7719 (1.0372) & 0.7964 (1.0701) & 0.8100 (1.0884) & 0.8339 (1.1205) \\
\midrule
\textbf{GloVe} \\
\midrule
Mean MC & 0.7397 & 0.7658 (1.0354) & 0.7853 (1.0354) & 0.8013 (\textbf{1.0833}) & 0.8202 (1.1088) \\
DE & 0.7397 & 0.7648 (1.0339) & \textbf{0.7940} (\textbf{1.0735}) & 0.7998 (1.0812) & 0.8204 (\textbf{1.1091}) \\
Softmax & \textbf{0.7442} & \textbf{0.7686} (\textbf{1.0328}) & 0.7918 (1.0639) & \textbf{0.8023} (1.0780) & \textbf{0.8217} (1.0141) \\
PL-Variance & \textbf{0.7442} & \textbf{0.7686} (\textbf{1.0328}) & 0.7918 (1.0639) & \textbf{0.8023} (1.0780) & 0.8204 (1.1023) \\ \bottomrule
\end{tabular}
} 
\caption{Macro F1 score and improvement rate for the SST-2 dataset.}
\label{table:performance_holdout_sst}
\end{table*}

\clearpage
\section{Model Calibration Plots}
\label{app:model_calibration}

\begin{figure*}[h]
    \begin{minipage}[t]{0.49\linewidth}
        \centering
        \includegraphics[width=1.0\linewidth]{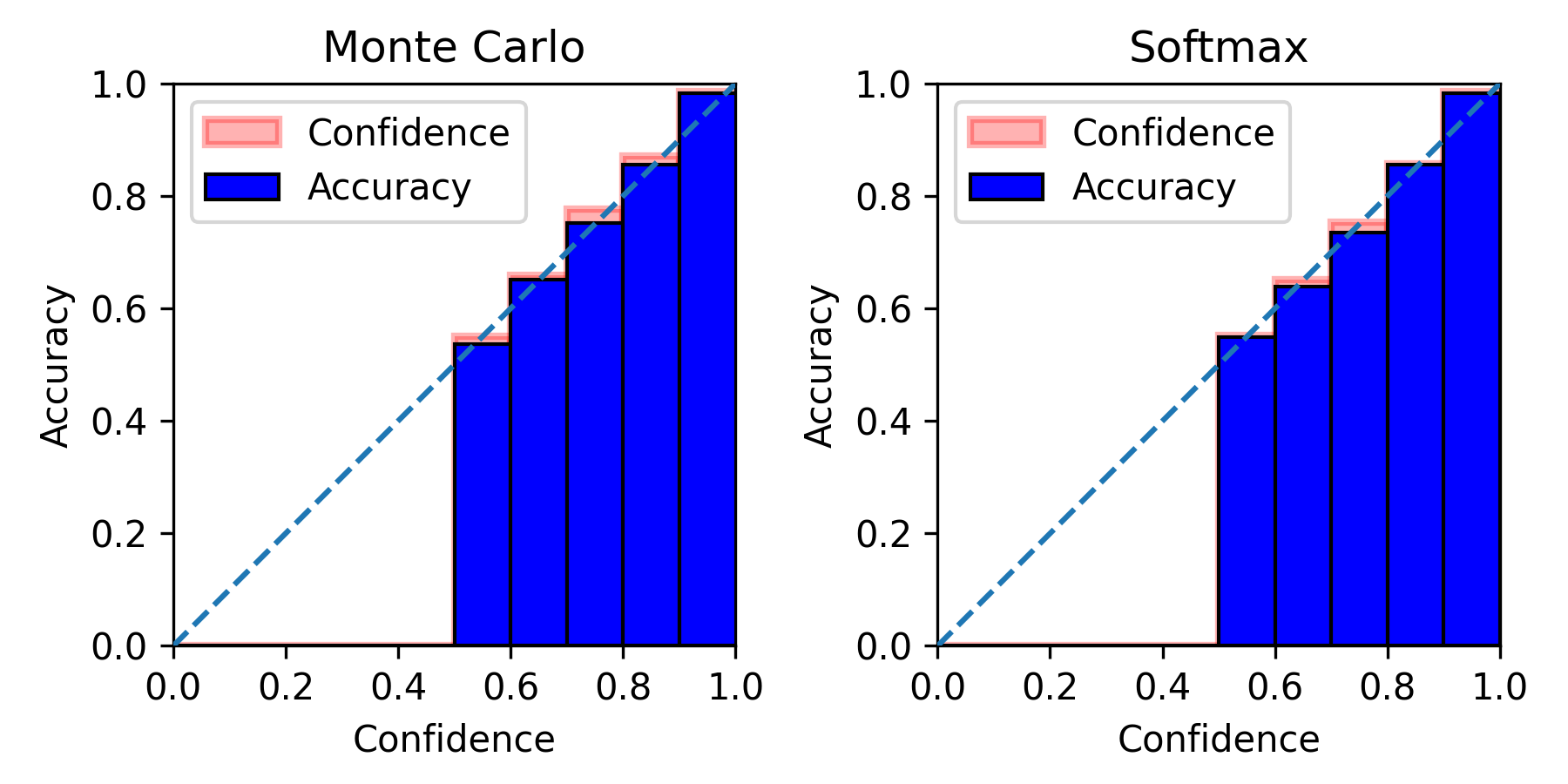}
    \end{minipage}
    \hfill
    \begin{minipage}[t]{0.49\linewidth}
        \centering
        \includegraphics[width=1.0\linewidth]{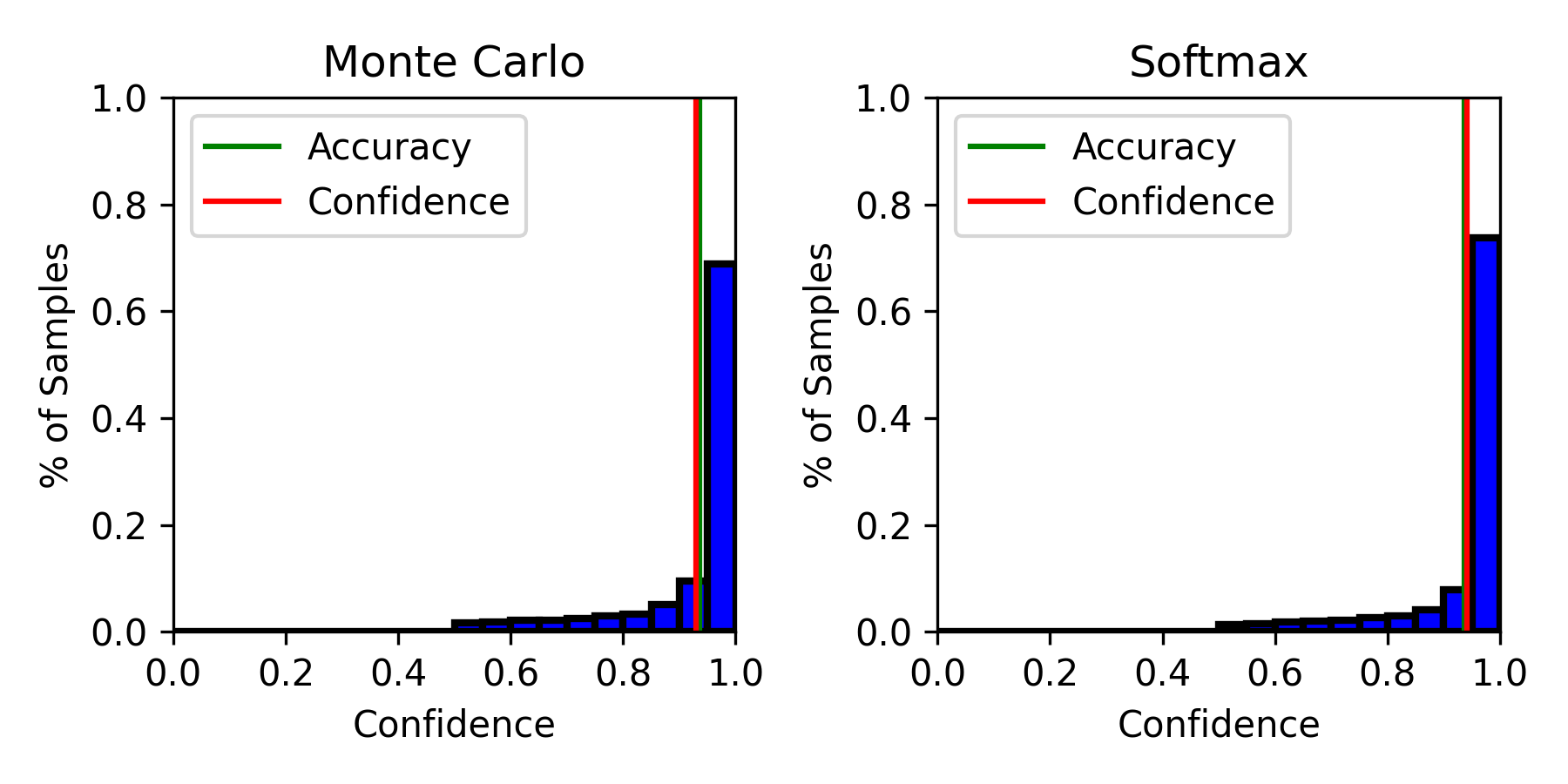}
    \end{minipage}
    \caption{Reliability diagram (left) and confidence histogram (right) of IMDb using BERT-CNN.}
    \label{fig:rel_conf_bert_imdb}
\end{figure*}

\begin{figure*}[!h]
    \begin{minipage}[t]{0.49\linewidth}
        \centering
        \includegraphics[width=1.0\linewidth]{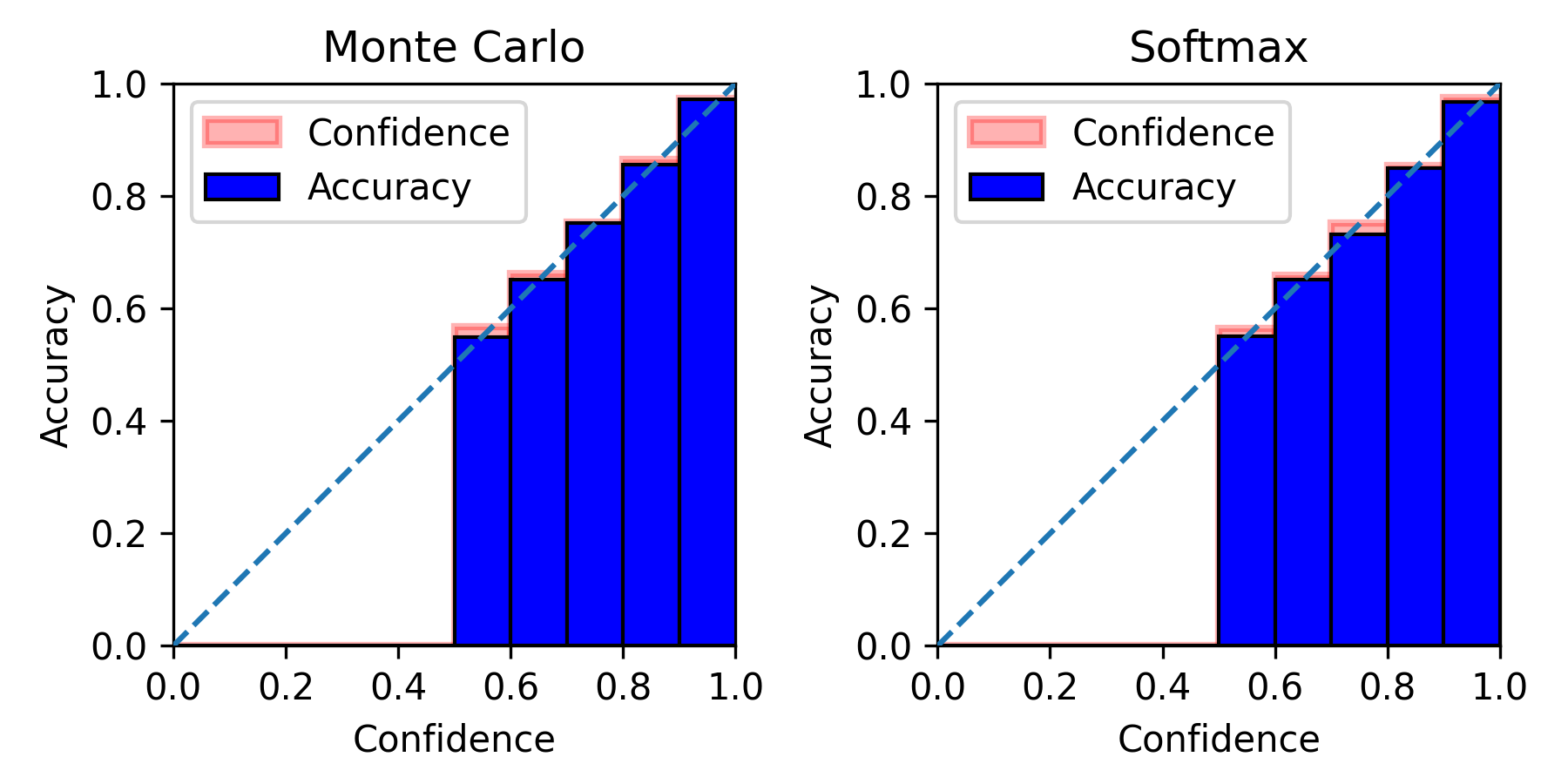}
    \end{minipage}
    \hfill
    \begin{minipage}[t]{0.49\linewidth}
        \centering
        \includegraphics[width=1.0\linewidth]{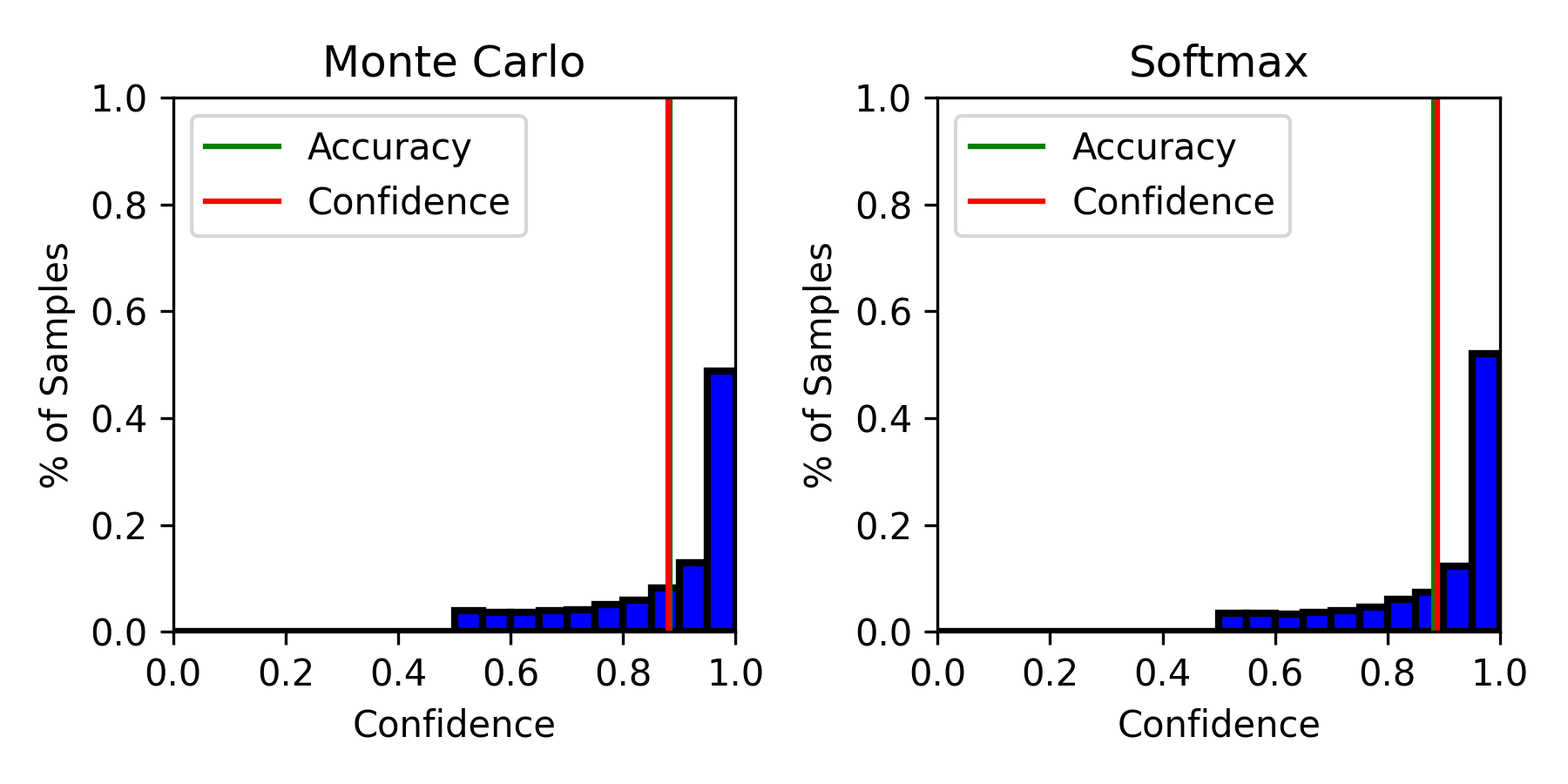}
    \end{minipage}
    \caption{Reliability diagram (left) and confidence histogram (right) of IMDb using GloVe-CNN.}
    \label{fig:rel_conf_glove_imdb}
\end{figure*}

\begin{figure*}[!h]
    \begin{minipage}[t]{0.49\linewidth}
        \centering
        \includegraphics[width=1.0\linewidth]{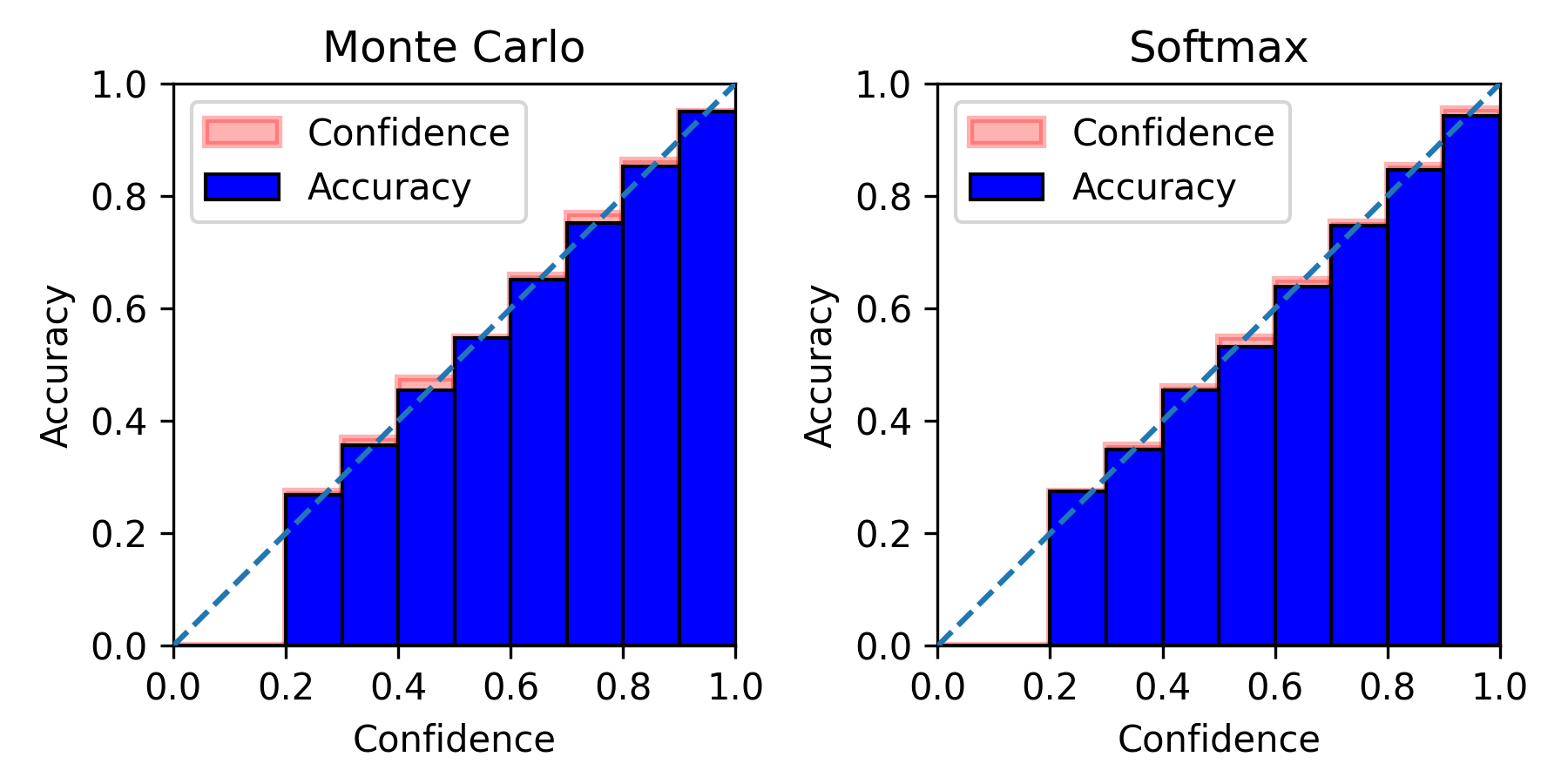}
    \end{minipage}
    \hfill
    \begin{minipage}[t]{0.49\linewidth}
        \centering
        \includegraphics[width=1.0\linewidth]{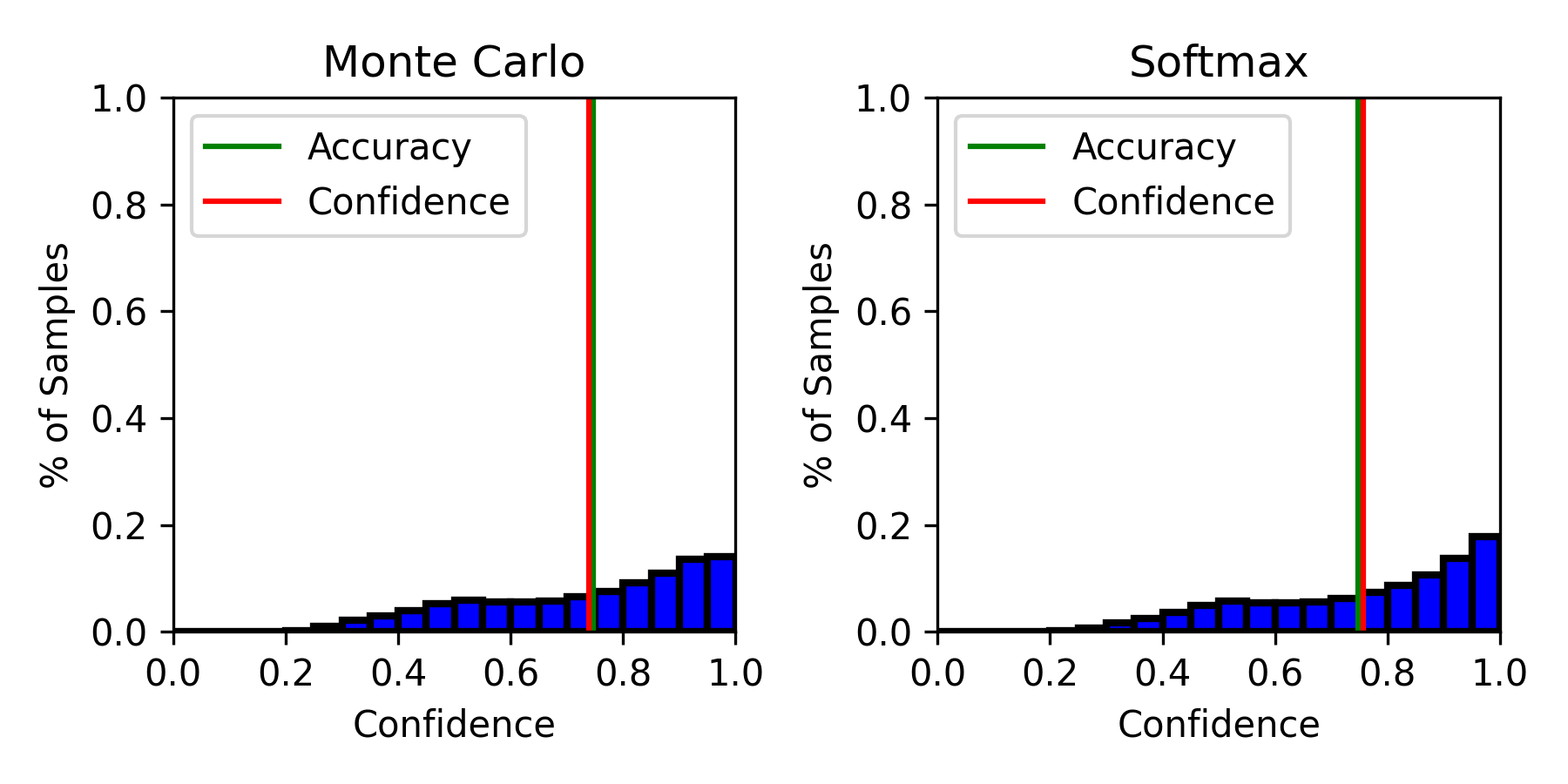}
    \end{minipage}
    \caption{Reliability diagram (left) and confidence histogram (right) of Amazon using BERT-CNN.}
    \label{fig:rel_conf_bert_amazon}
\end{figure*}

\begin{figure*}[!h]
    \begin{minipage}[t]{0.49\linewidth}
        \centering
        \includegraphics[width=1.0\linewidth]{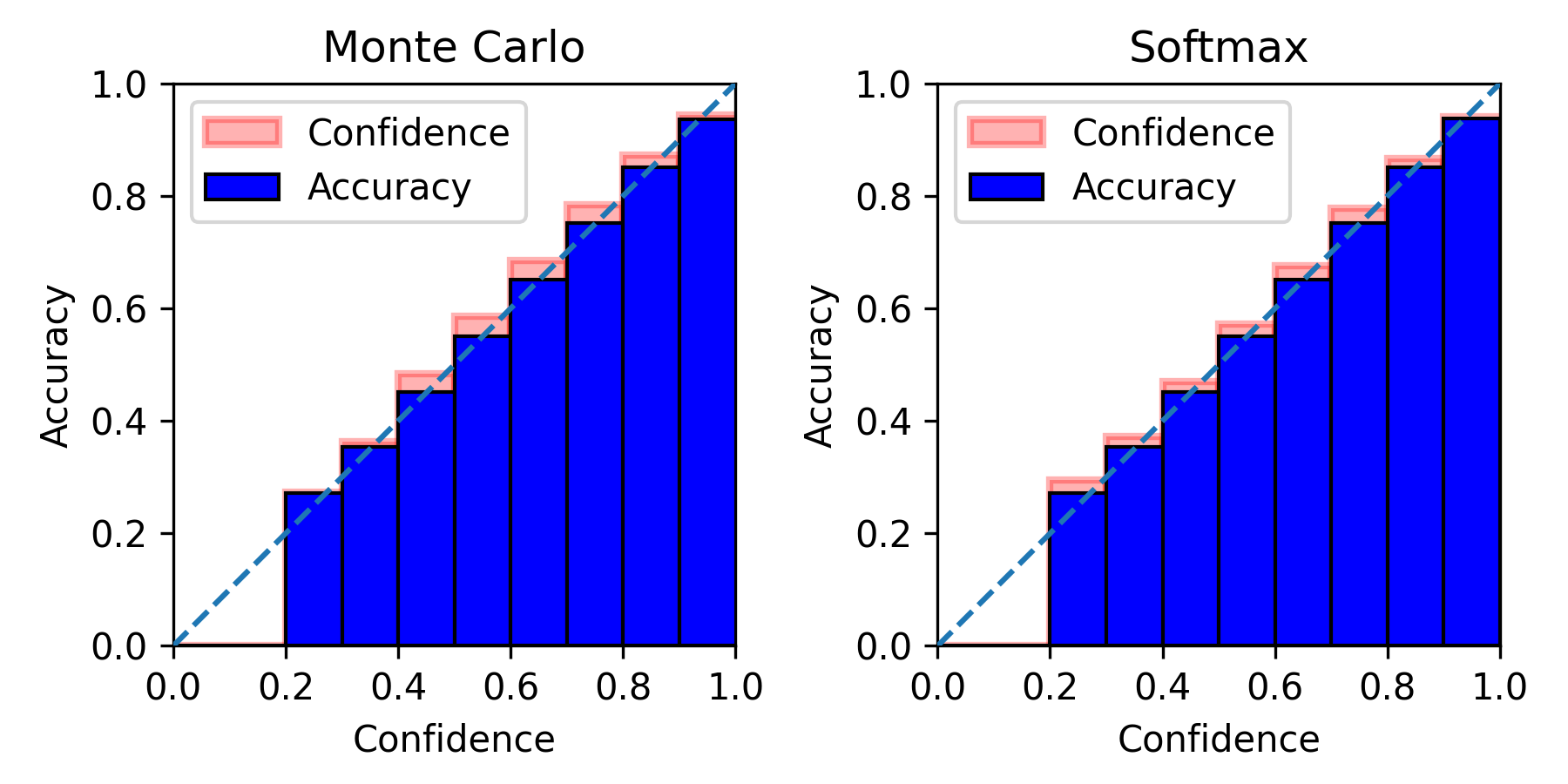}
    \end{minipage}
    \hfill
    \begin{minipage}[t]{0.49\linewidth}
        \centering
        \includegraphics[width=1.0\linewidth]{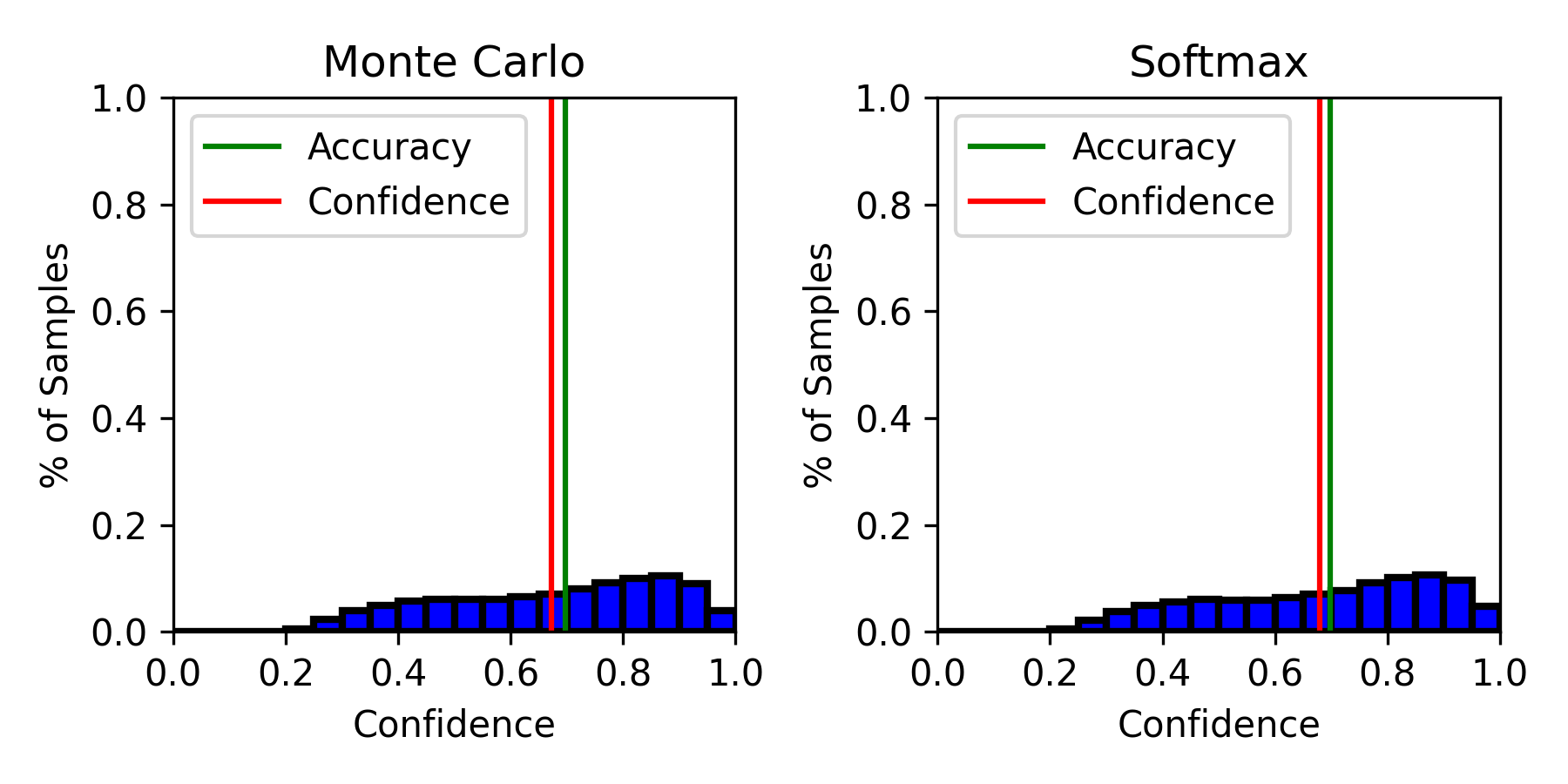}
    \end{minipage}
    \caption{Reliability diagram (left) and confidence histogram (right) of Amazon using GloVe-CNN.}
    \label{fig:rel_conf_glove_amazon}
\end{figure*}

\begin{figure*}[h]
    \begin{minipage}[t]{0.49\linewidth}
        \centering
        \includegraphics[width=1.0\linewidth]{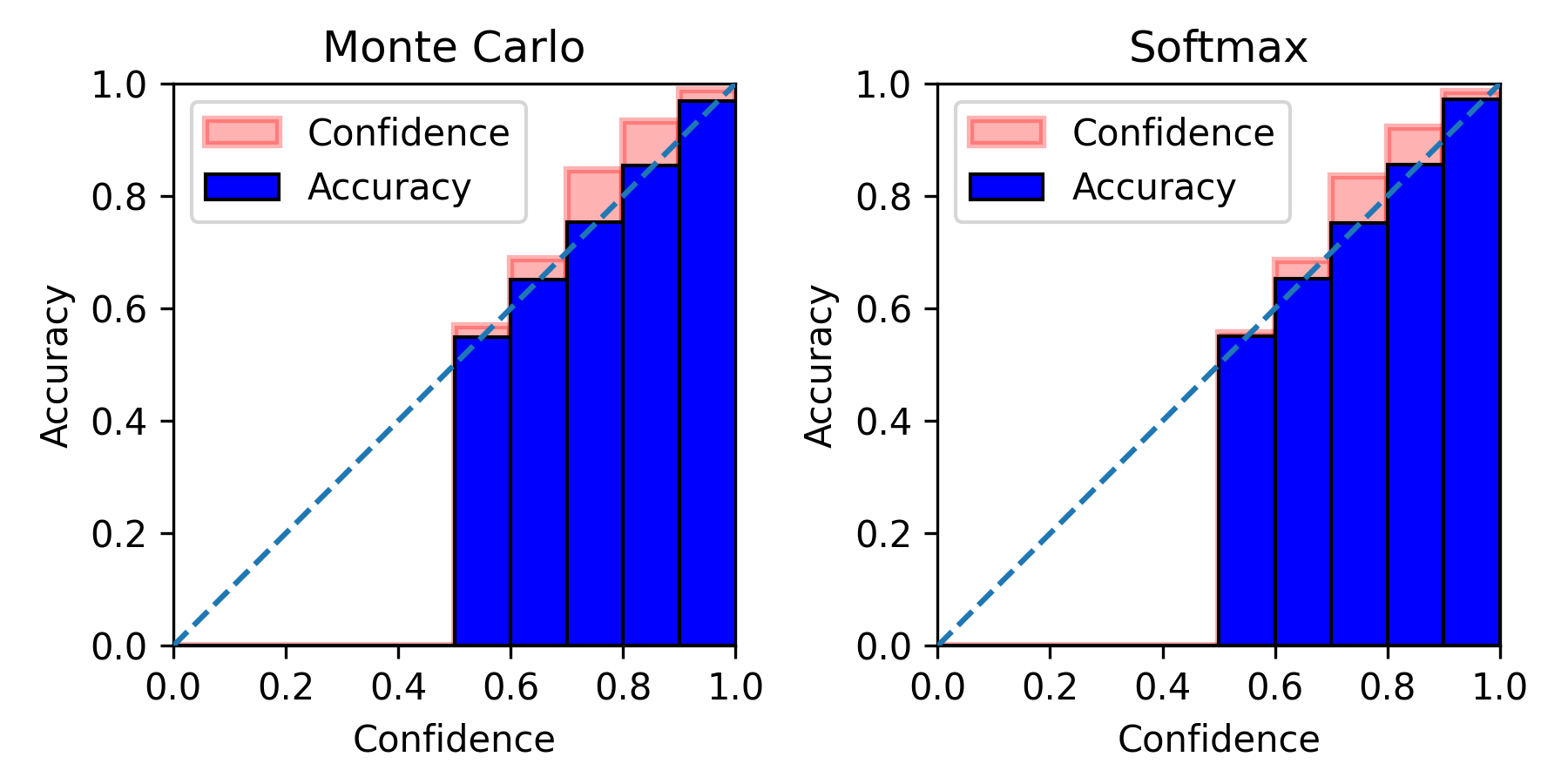}
    \end{minipage}
    \hfill
    \begin{minipage}[t]{0.49\linewidth}
        \centering
        \includegraphics[width=1.0\linewidth]{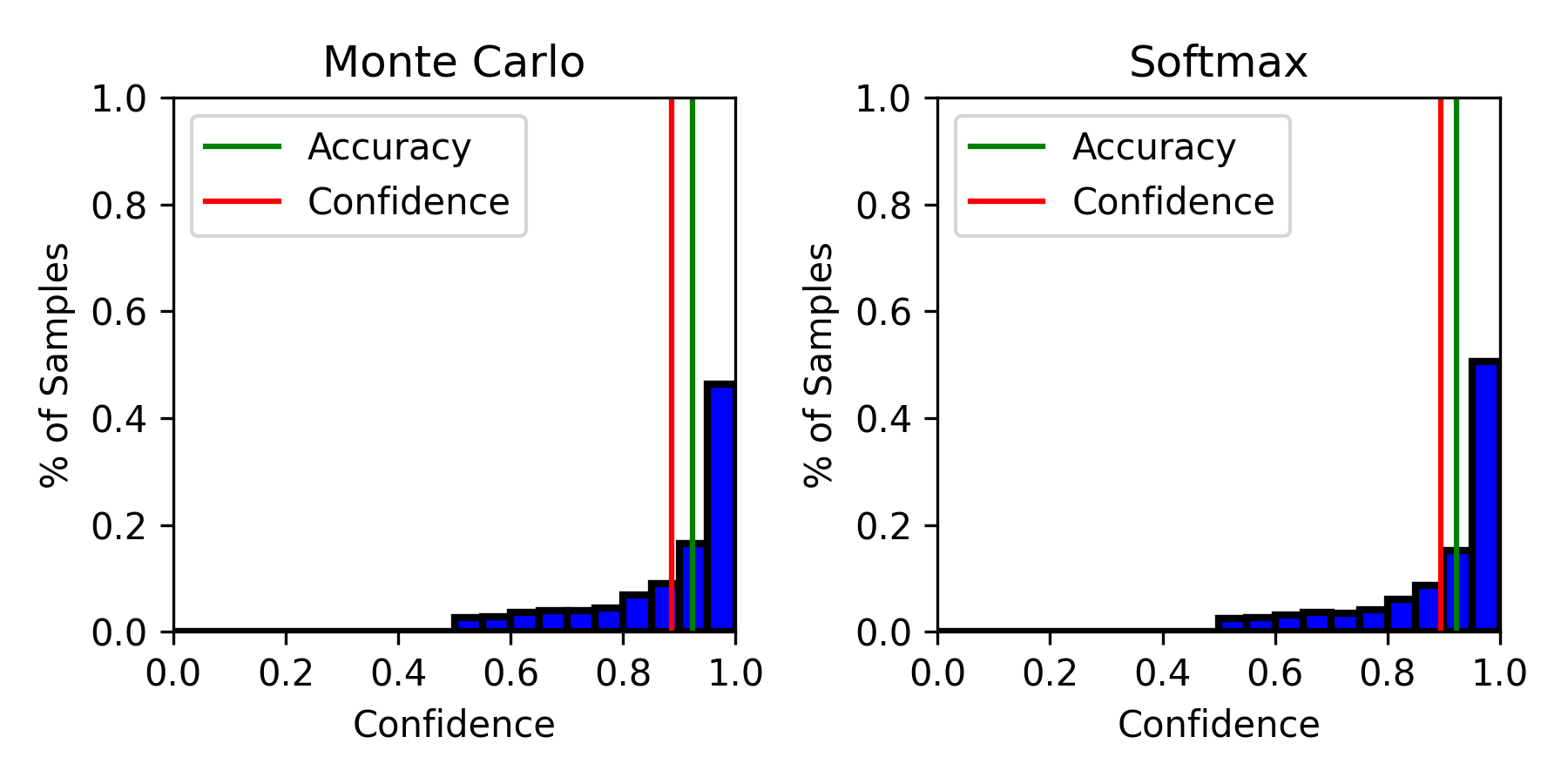}
    \end{minipage}
    \caption{Reliability diagram (left) and confidence histogram (right) of WIKI using BERT-CNN.}
    \label{fig:rel_conf_bert_wiki}
\end{figure*}

\begin{figure*}[!h]
    \begin{minipage}[t]{0.49\linewidth}
        \centering
        \includegraphics[width=1.0\linewidth]{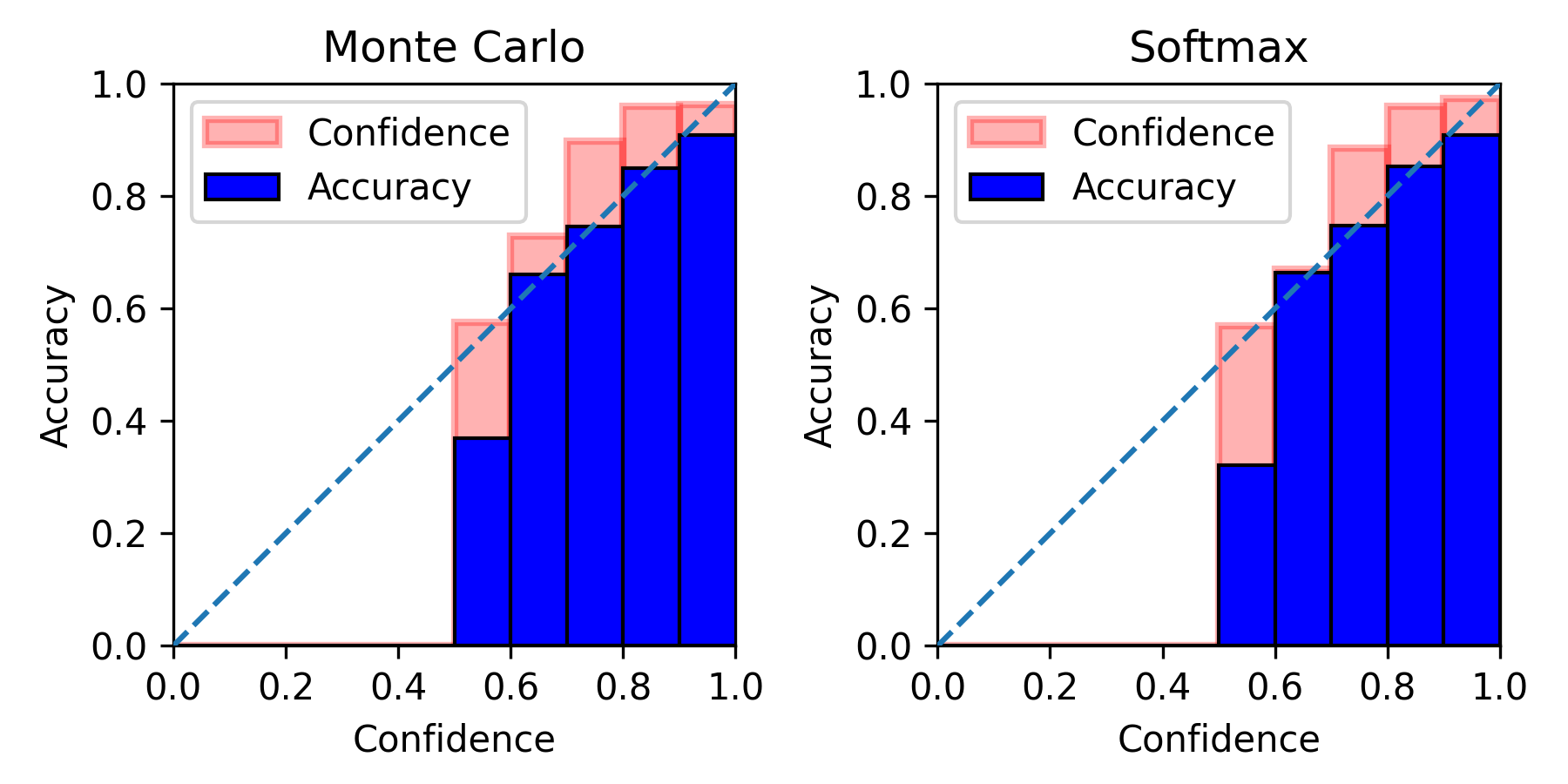}
    \end{minipage}
    \hfill
    \begin{minipage}[t]{0.49\linewidth}
        \centering
        \includegraphics[width=1.0\linewidth]{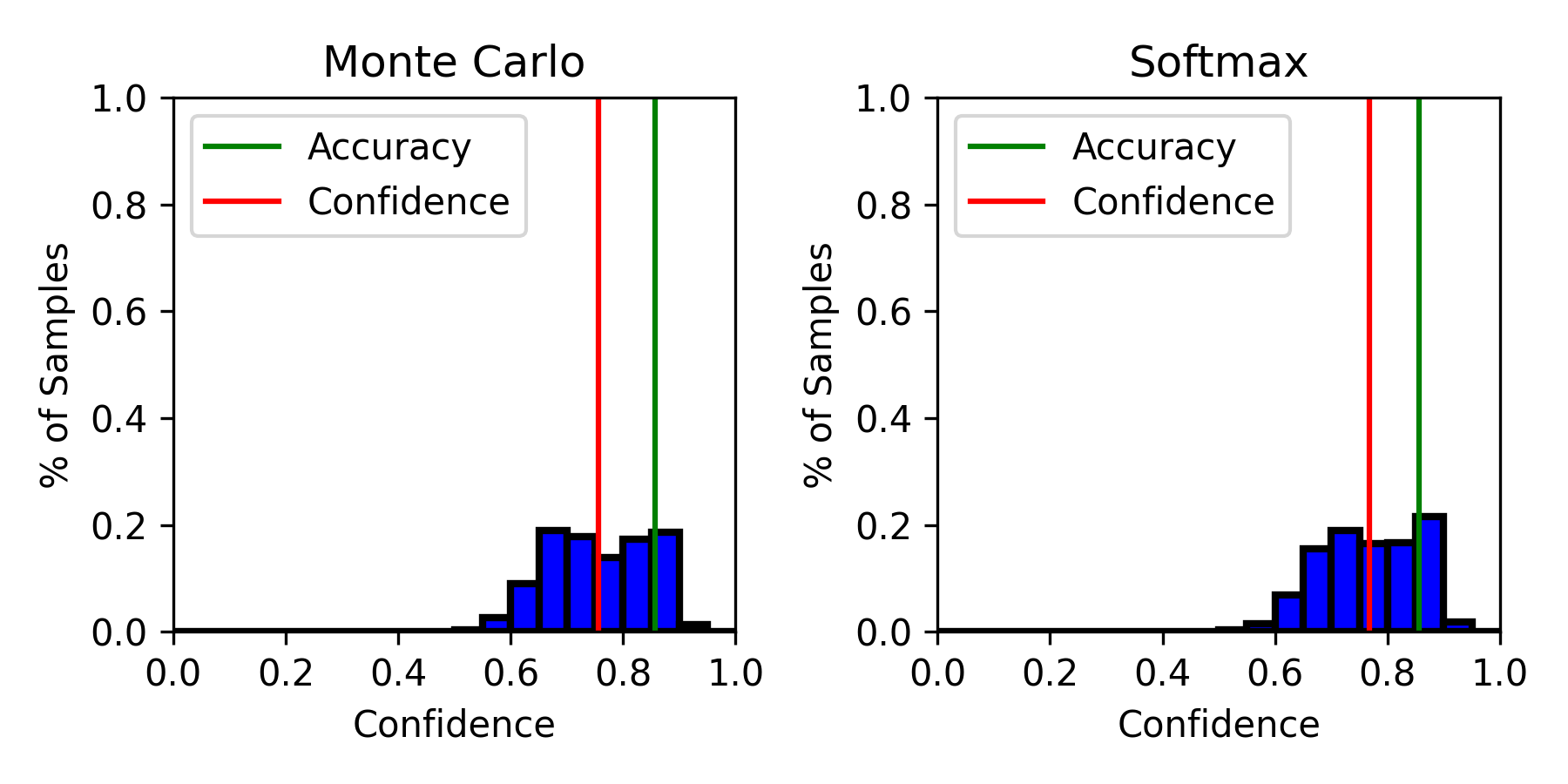}
    \end{minipage}
    \caption{Reliability diagram (left) and confidence histogram (right) of WIKI using GloVe-CNN.}
    \label{fig:rel_conf_glove_wiki}
\end{figure*}

\begin{figure*}[h]
    \begin{minipage}[t]{0.49\linewidth}
        \centering
        \includegraphics[width=1.0\linewidth]{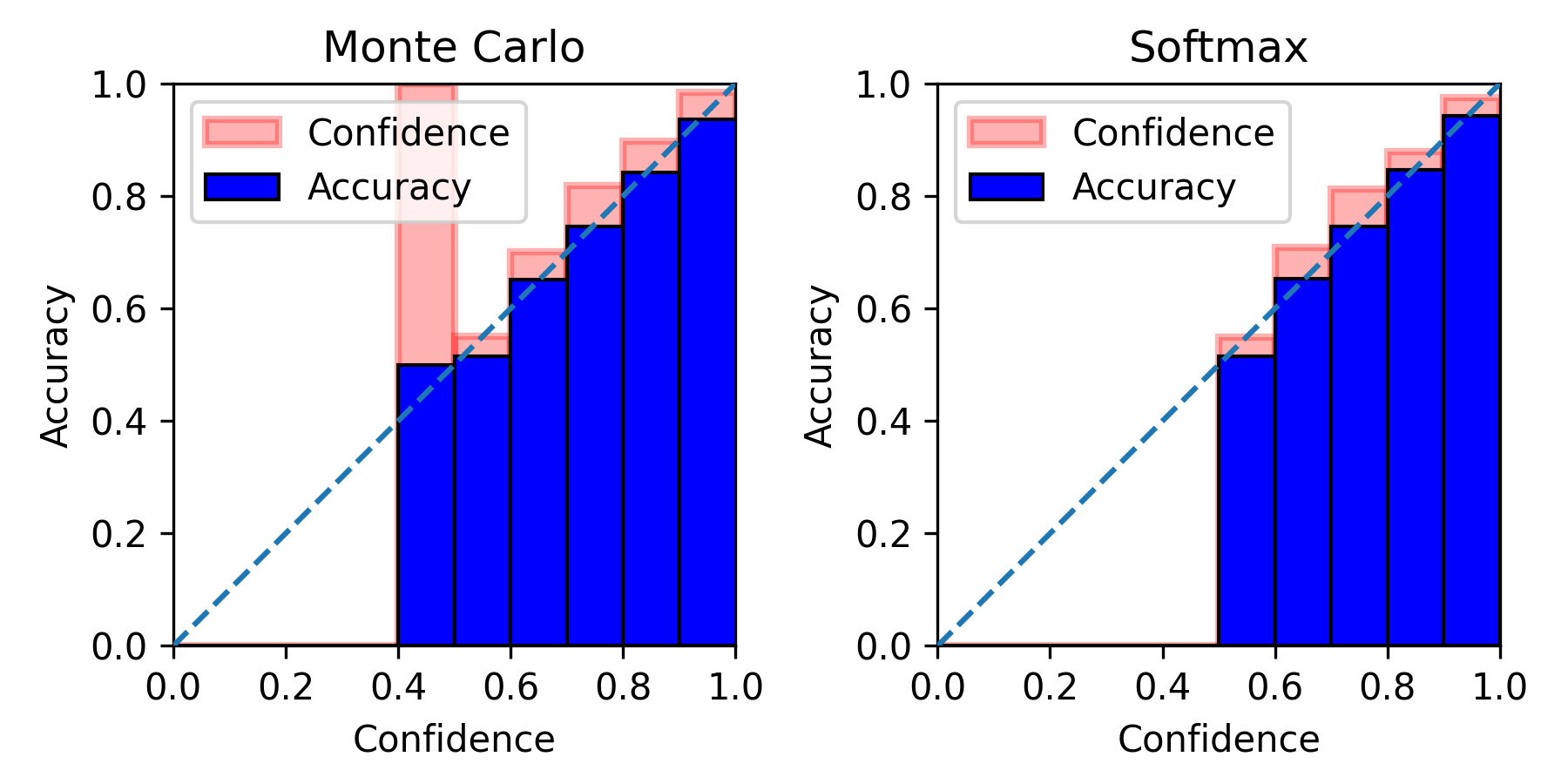}
    \end{minipage}
    \hfill
    \begin{minipage}[t]{0.49\linewidth}
        \centering
        \includegraphics[width=1.0\linewidth]{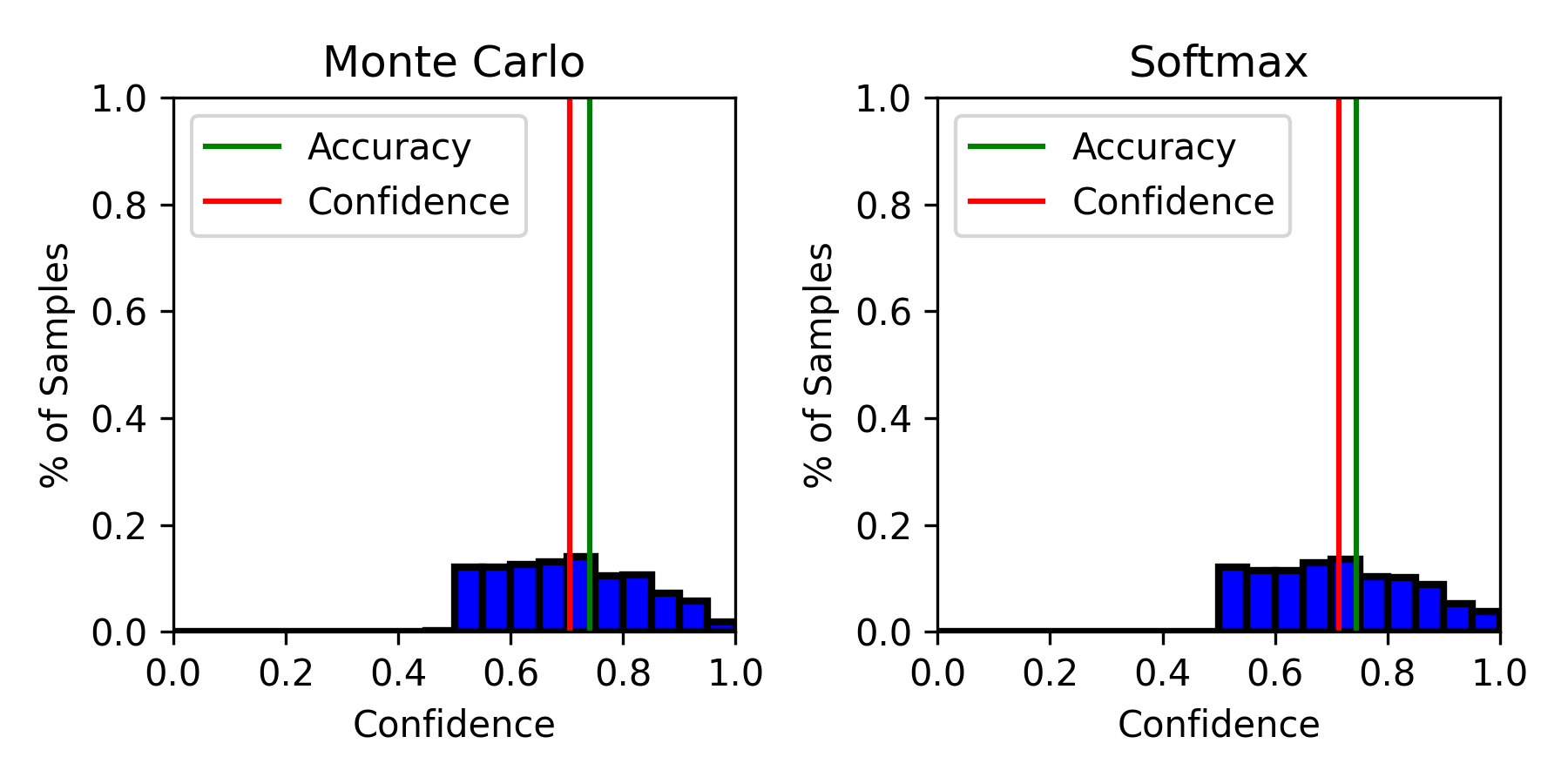}
    \end{minipage}
    \caption{Reliability diagram (left) and confidence histogram (right) of SST-2 using BERT-CNN.}
    \label{fig:rel_conf_bert_wiki}
\end{figure*}

\begin{figure*}[!h]
    \begin{minipage}[t]{0.49\linewidth}
        \centering
        \includegraphics[width=1.0\linewidth]{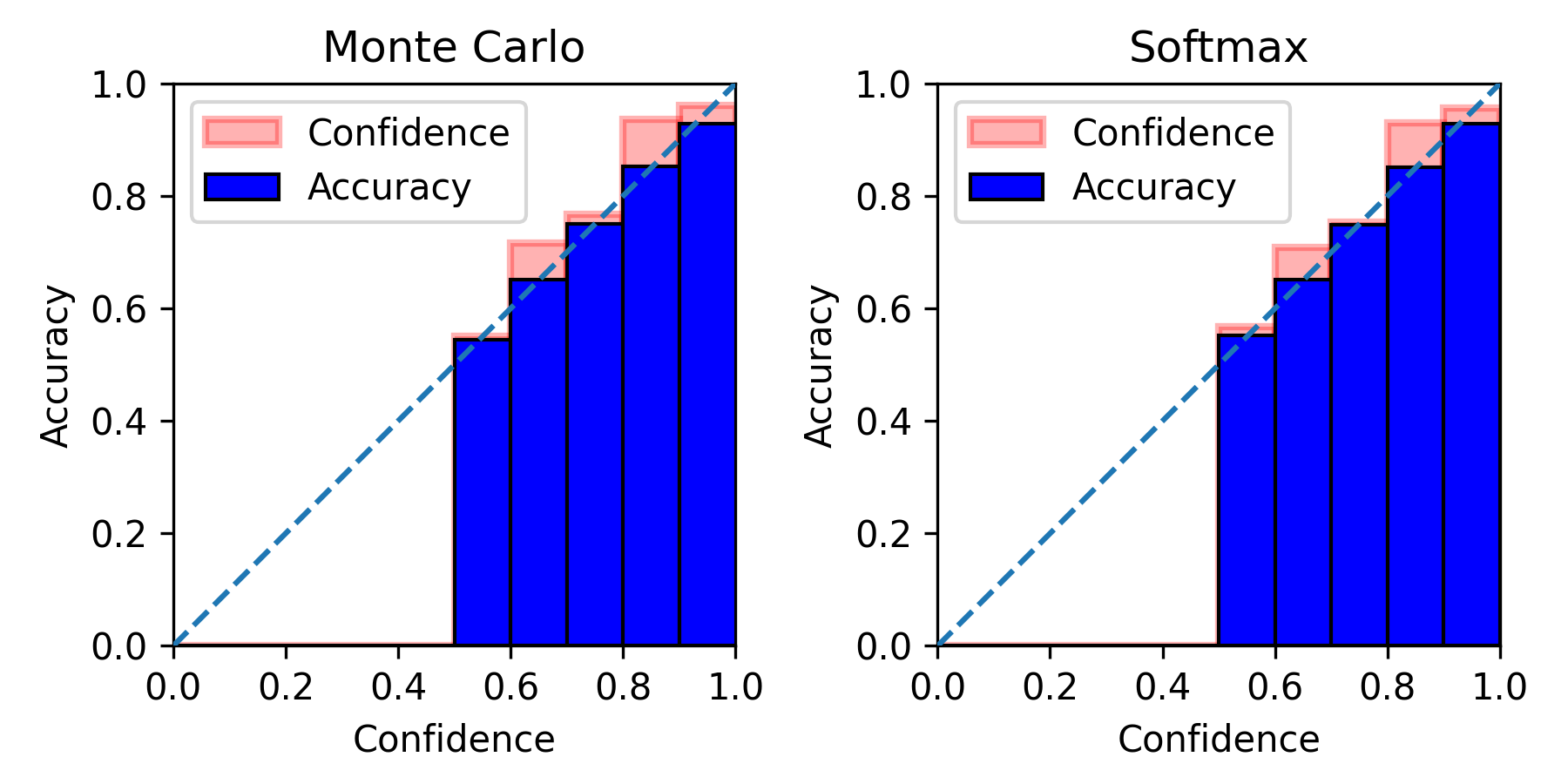}
    \end{minipage}
    \hfill
    \begin{minipage}[t]{0.49\linewidth}
        \centering
        \includegraphics[width=1.0\linewidth]{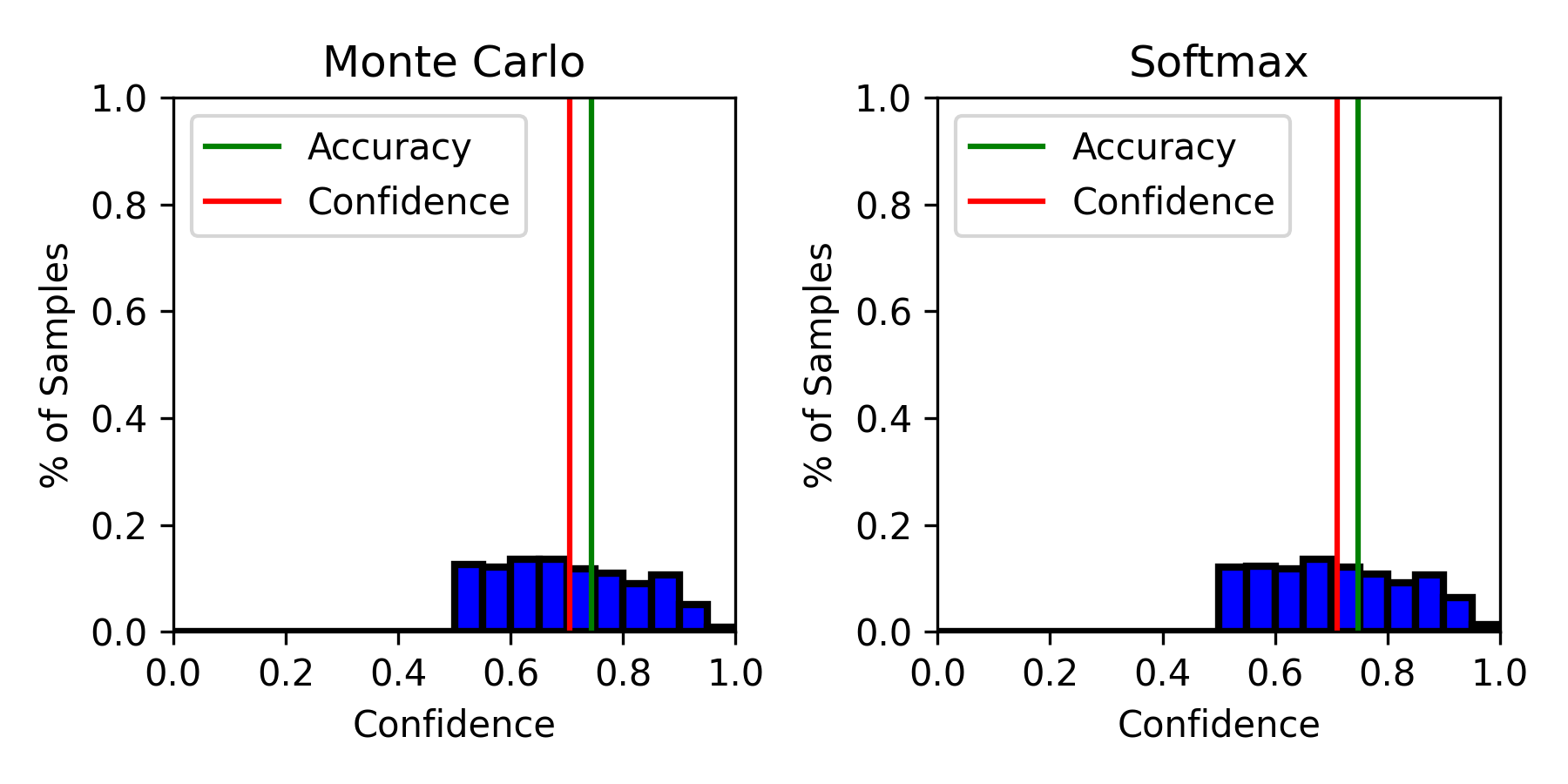}
    \end{minipage}
    \caption{Reliability diagram (left) and confidence histogram (right) of SST-2 using GloVe-CNN.}
    \label{fig:rel_conf_glove_wiki}
\end{figure*}

\end{document}